\documentclass[manuscript,screen,authorversion]{acmart}
\AtBeginDocument{%
  }
    
\setcopyright{acmlicensed}
\copyrightyear{2018}
\acmYear{2018}
\acmDOI{XXXXXXX.XXXXXXX}
\acmConference[Conference acronym 'XX]{Make sure to enter the correct
  conference title from your rights confirmation email}{June 03--05,
  2018}{Woodstock, NY}
\acmISBN{978-1-4503-XXXX-X/2018/06}

\usepackage{amsmath}
\usepackage{amsthm}
\usepackage{booktabs}
\usepackage{framed} 
\usepackage{multirow}
\usepackage{booktabs}
\usepackage{enumitem}
\usepackage{caption}
\usepackage{graphicx}
\usepackage{float} 
\usepackage{etoolbox}
\usepackage{xspace}

\usepackage{color}
\usepackage{subfigure}

\newtheorem{problem}{Problem}
\newtheorem{definition}{Definition}
\newtheorem{prop}{Proposition}

\newcommand{\eg}{\emph{e.g.},\xspace}

\newcommand{\ie}{\emph{i.e.},\xspace}

\newcommand{\eat}[1]{}

\begin{document}

\title{Unsupervised Graph Anomaly Detection via Multi-Hypersphere Heterophilic Graph Learning}

\author{HANG NI}
\affiliation{%
  \institution{The Hong Kong University of Science and Technology (Guangzhou)}
  \city{Guangzhou}
  \country{China}}
\email{hni017@connect.hkust-gz.edu.cn}

\author{JINDONG HAN}
\affiliation{%
  \institution{The Hong Kong University of Science and Technology}
  \city{Hong Kong}
  \country{China}}
\email{jhanao@connect.ust.hk}

\author{NENGJUN ZHU}
\affiliation{%
  \institution{Shanghai University}
  \city{Shanghai}
  \country{China}}
\email{zhu\_nj@shu.edu.cn}

\author{HAO LIU}
\affiliation{%
  \institution{The Hong Kong University of Science and Technology (Guangzhou)}
  \city{Guangzhou}
  \country{China}}
\email{liuh@ust.hk}

\renewcommand{\shortauthors}{Ni et al.}

\begin{abstract}
\eat{Graph anomaly detection (GAD) aims to identify the nodes that deviate from the majorities, which is a key data mining task with diverse Web applications, ranging from financial fraud detection to malicious user detection. Since the lack of ground truths, unsupervised graph anomaly detection (UGAD) has received much attention. However, most existing methods follow a two-stage framework which lacks sufficient distinguishing ability for graph anomaly detection. Specifically, they fail to identify anomalies camouflaged in vast normal neighbors or hidden in local contexts of communities. To address these problems, we propose an anomaly-distinguishing framework \textsl{AnoDIS} for unsupervised graph anomaly detection. Firstly, we propose \textsl{Anomaly-aware Heterophilic Graph Neural Network (AnoHetGNN)} to learn representations in an unsupervised manner. Secondly, taking into count anomalies from local contexts, we present a \textsl{Multi-hypersphere Learning (MHL)} module for anomaly identification, via contrastive-based community detection. Besides, an \textsl{Embedding-aware Contrastive Learning (ECL)} paradigm is designed for community detection to ensure the robust training of multi-hypersphere learning. Finally, we conduct extensive experiments on various datasets which demonstrate the superiority of our method.}
Graph Anomaly Detection (GAD) plays a vital role in various data mining applications such as e-commerce fraud prevention and malicious user detection. Recently, Graph Neural Network (GNN) based approach has demonstrated great effectiveness in GAD by first encoding graph data into low-dimensional representations and then identifying anomalies under the guidance of supervised or unsupervised signals. However, existing GNN-based approaches implicitly follow the homophily principle (\ie the "like attracts like" phenomenon) and fail to learn discriminative embedding for anomalies that connect vast normal nodes. Moreover, such approaches identify anomalies in a unified global perspective but overlook diversified abnormal patterns conditioned on local graph context, leading to suboptimal performance. To overcome the aforementioned limitations, in this paper, we propose a \emph{\textbf{M}ulti-hypersphere \textbf{Het}erophilic \textbf{G}raph \textbf{L}earning}~(\textbf{MHetGL}) framework for unsupervised GAD. Specifically, we first devise a \textsl{Heterophilic Graph Encoding} (HGE) module to learn distinguishable representations for potential anomalies by purifying and augmenting their neighborhood in a fully unsupervised manner. Then, we propose a \textsl{Multi-Hypersphere Learning} (MHL) module to enhance the detection capability for context-dependent anomalies by jointly incorporating critical patterns from both global and local perspectives. Extensive experiments on ten real-world datasets show that MHetGL outperforms 14 baselines. Our code is publicly available at \url{https://github.com/KennyNH/MHetGL}.
\end{abstract}

\ccsdesc[500]{Information systems~Data mining}

\keywords{Unsupervised graph anomaly detection, graph neural network, hypersphere learning}

\received{20 February 2007}
\received[revised]{12 March 2009}
\received[accepted]{5 June 2009}

\maketitle

\section{Introduction}
Graph Anomaly Detection (GAD) aims to identify anomalous graph objects (\eg nodes, edges, subgraphs) that deviate significantly from the majority in graph-structured data \cite{kim2022graph}, which plays a pivotal role in various applications, such as preventing e-commerce fraudulent activities~\cite{zhang2022efraudcom} and detecting malicious users~\cite{gao2017anomaly}. 
However, due to the intricate characteristics and rare occurrence of anomalies in graph data, manually labeling them can be time-consuming and error-prone~\cite{ma2021comprehensive}. 
Therefore, in recent years, unsupervised GAD has attracted significant attention from both academia~\cite{liu2022benchmarking} and industry~\cite{chen2022antibenford}. 

Recently, extensive efforts have been made to tackle unsupervised GAD by utilizing advanced Graph Neural Network (GNN) techniques~\cite{kim2022graph,liu2022benchmarking,liu2021anomaly,ding2019deep,wang2021one}. 
These approaches share a general framework that typically consists of two stages: (1) \emph{representation learning}, and (2) \emph{anomaly identification}. 
The first stage aims to learn discriminative node representations from complex graph data by simultaneously capturing node attributes and structural semantics~\cite{ding2019deep,fan2020anomalydae,wang2021one}. 
The second stage aims to identify anomalous node representations based on pre-defined unsupervised schemes, such as feature reconstruction~\cite{ding2019deep,roy2023gad}, contrastive learning~\cite{liu2021anomaly,duan2023graph}, and one-class classification~\cite{wang2021one,zhou2021subtractive}. 
To name a few, DOMINANT~\cite{ding2019deep} 
leverages the reconstruction errors of node features as anomaly scores, whereas OCGNN~\cite{wang2021one} spots anomalies by measuring the distance of nodes to a predefined reference point in an anomaly-aware vector space. 
Despite the fruitful progress made so far, two major challenges significantly hinder the effectiveness of existing unsupervised GAD approaches.
We detail each of them below.


\textsl{(1) Homophily-induced indistinguishability.} 
In the representation learning stage, existing GNN-based approaches derive node embeddings by aggregating and transforming neighborhood information.
Recent studies~\cite{zhu2021graph,zheng2022graph} have uncovered that the success of GNNs can be attributed to the homophily principle, \ie nodes with the same class are more likely to be connected, which provides additional information to enhance the original node representations during the message passing process. 
However, graph anomalies are usually extremely sparse and exhibit strong heterophily~\cite{dou2020enhancing}, which means the connected neighbors possess different properties or classes. 
For example, crafty fraudsters can camouflage themselves by establishing connections with benign users to escape detection. 
When GNN recursively aggregates information from neighboring normal nodes, the anomaly signals will be smoothed and diluted, making them indistinguishable. 
A few studies have attempted to overcome the limitation of the heterophily issue via separate aggregation functions~\cite{liu2021pick,shi2022h2,gao2023addressing} or frequency-dependent signal channels~\cite{liu2021pick,shi2022h2,gao2023addressing}. 
However, these approaches either rely on abundant labels for supervision or suffer from class imbalance issues in anomalous node detection, which are not applicable to graph anomaly detection in a fully unsupervised way.
Therefore, how to tackle the heterophily problem to improve unsupervised graph anomaly distinguishability is the first challenge.

\textsl{(2) Uniformity-induced indistinguishability.} 
In the anomaly identification stage, existing approaches seamlessly apply uniform criteria to detect anomalous nodes but overlook abnormal behaviors from the local perspective.
For instance, one-class classification methods~\cite{wang2021one,zhou2021subtractive} employ hypersphere learning that tries to uniformly enclose the representations of all the normal nodes within a hypersphere and regards the nodes outside the hypersphere as anomalies. 
However, the real-world anomalies can be highly context-dependent, \eg conditioned on local communities~\cite{gao2010community,zhou2021subtractive}. Take the financial network as an example~\cite{huang2022dgraph}, a node involved in frequent large transactions could be considered anomalous in low-income communities but deemed normal in high-income communities. The uniform identification schemes fall short of identifying such anomalies that rely on the local context, resulting in suboptimal detection performance. 
Therefore, how to collectively identify anomalies from both global and local perspectives for more effective unsupervised GAD is another challenge.

To bridge the aforementioned gaps, in this paper, we propose a \emph{\textbf{M}ulti-hypersphere \textbf{Het}erophilic \textbf{G}raph \textbf{L}earning}~(\textbf{MHetGL}) framework for unsupervised GAD. 
Specifically, we first devise a \textsl{Heterophilic Graph Encoding} (HGE) module to learn distinguishable representations by adaptively selecting informative neighboring nodes for message passing and aggregation. 
In particular, we develop a training-free homophily-guided neighborhood refinement block, which manipulates the graph topology by purifying and augmenting the homophilic neighborhood for each node. 
Besides, we design an anomaly-aware aggregation block to perform message passing on the manipulated graph structure, which preserves critical anomaly knowledge to address the first challenge. 
Then, based on the representations derived from HGE, we propose a \textsl{Multi-Hypersphere Learning} (MHL) module to enhance the context-dependent anomaly distinguishability.
By enclosing both global and local specific normal patterns within multiple learnable hyperspheres in the latent space, MHL further improves the capability of detecting context-dependent anomalies, thereby solving the second challenge. Moreover, a hypersphere regularization block is devised to avoid model collapse when optimizing the hypersphere objective~\cite{qiu2022raising}.
Our major contributions are summarized as follows:
\begin{itemize}[leftmargin=*]
\item {We devise a Heterophilic Graph Encoding (HGE) module to enhance the effectiveness of message passing on heterophilous graph nodes and derive more distinguishable representations for GAD in a fully unsupervised manner.} \ \ \  

\item {We propose a Multi-Hypersphere Learning (MHL) module to identify context-dependent anomalies by jointly incorporating critical patterns from both global and local perspectives. A tailored hypersphere regularization objective is further introduced to stabilize the learning process.} \ \ \  

\item {We conduct extensive experiments on ten real-world datasets, and the results validate that our method can significantly improve the performance of unsupervised GAD.}
\end{itemize}
\section{Related Work}
In this section, we review related works including graph neural networks and unsupervised graph anomaly detection.

\noindent\textbf{Graph neural network.} 
Graph Neural Networks (GNNs) have achieved great success in transforming relational graph data into informative representations, including spectral GNNs (\eg GCN~\cite{gcn}) and spatial GNNs (\eg GraphSAGE~\cite{graphsage}). The effectiveness of these models hinge on the homophily principle of graph data, \ie the connected nodes are prone to sharing the same class.
However, nodes with different classes may be linked in the real-world graph data, which may negatively affect the performance of vanilla GNNs following the homophily assumption. Recently, tremendous efforts have been devoted to developing heterophilic GNNs~\cite{zheng2022graph}. Specifically, these works either try to discriminate neighbors with different classes, as the uniform aggregation ignores the distinction of information between similar and dissimilar neighbors, or try to discover latent homophilic neighbors, as the local aggregation paradigm fails to exploit informative nodes far apart~\cite{zheng2022graph}. However, these methods are restricted in semi-supervised node classification tasks and heavily rely on label information~\cite{zheng2022graph}. 

Recently, there have arisen some preliminary works~\cite{liu2023beyond,lin2023multi,xiao2022decoupled} of unsupervised representation learning on graphs with heterophily. GREET~\cite{liu2023beyond} designs an explicit edge discriminator and proposes a pivot-anchored ranking loss to train the discriminating module in an unsupervised manner. MVGE~\cite{lin2023multi} utilizes ego and walk-based aggregated features for reconstruction, to respectively filter high-frequency and low-frequency signals. DSSL~\cite{xiao2022decoupled} assumes each node has latent heterogeneous factors that are utilized to make connections to its different neighbors and models the neighborhood distribution via a mixture of generative processes in the representation space. However, these unsupervised heterophilic graph learning methods are used for general graph tasks and are difficult to directly apply to GAD due to the complicated characteristics of graph anomalies such as class imbalance.
In this work, we devise a heterophilic graph encoding module tailored for GAD that can address the above problem.





\noindent\textbf{Unsupervised graph anomaly detection.} 
In the past decade, various approaches have been proposed for unsupervised GAD. Reconstruction-based methods~\cite{ding2019deep,fan2020anomalydae,huang2023unsupervised,roy2023gad} usually adopt autoencoder or GAN as the backbone, which aims to reconstruct the structural or contextual information of raw graph data. After model training, the objects with higher reconstruction errors are defined as anomalies. Contrastive-based methods~\cite{liu2021anomaly,zheng2022unsupervised,duan2023graph} assume the anomalies are different from its ego-subgraph and use the contrastive objective as a constraint. Hypersphere-based methods~\cite{wang2021one,zhou2021subtractive} uses hypersphere learning, which constrains all the normal objects with a hypersphere in the embedding space and representations outside the hypersphere will be identified as anomalies.

Recently, hypersphere learning has received considerable attention in unsupervised GAD. To name a few, OCGNN~\cite{wang2021one} uses GNNs for node representations and a hypersphere objective for optimization and anomaly scoring. To address the collapse problem in hypersphere learning, OCGTL~\cite{qiu2022raising} designs a neural transformation learning module that uses multiple GNNs to learn representations jointly. However, this method causes tedious manual trials for weight tuning and extra model complexity, which makes it hard to scale to multi-hypersphere learning scenarios. To detect unseen types of anomalies, MHGL~\cite{zhou2022unseen} proposes a multiple hypersphere learning module.
In this work, we propose a multi-hypersphere learning module for anomaly identification, which considers both global and local perspectives. Compared to MHGL, we automatically learn the local hypersphere center via a contrastive-based community detection module and retain the global perspective for anomaly identification. Besides, we propose a tailored hypersphere regularization block to alleviate the collapse problem in multi-hypersphere learning efficiently.

\section{Preliminary}

In this section, we first present some notations and definitions. Then we introduce the problem setting of unsupervised GAD in this work.

\begin{definition}
\textbf{Attributed Graph}. An attributed graph is defined as $\mathcal{G}=(\mathcal{V},\mathcal{E},\mathbf{X})$ where $\mathcal{V}=\{v_1,v_2,...,v_N\}$ denotes the set of nodes, $e_{ij} \in \mathcal{E}$ is the edge between node $v_i$ and $v_j$, and $\mathbf{X}\in \mathbb{R}^{N\times d_{in}}$ represents the attribute matrix. The i-$th$ row vector $\mathbf{x}_i\in \mathbb{R}^{d_{in}}$ of $\mathbf{X}$ denotes the attributes of node $v_i$. We also define the adjacency matrix of the attribute graph as $\mathbf{A}$.
\end{definition}

\begin{problem}
Given an attribute graph $\mathcal{G}=(\mathcal{V},\mathcal{E},\mathbf{X})$, we aim to learn an anomaly scoring model $s(\cdot)$ by leveraging unlabeled training data $\mathcal{V}^{train}$. The learned model can be utilized to predict anomaly scores for nodes in the testing data $\mathcal{V}^{test}$, and the nodes with the highest anomaly scores are labeled as anomalies.
\end{problem}

Note in this paper, we focus on node-level anomaly detection under the unsupervised setting. This is a more challenging setting than previous supervised GAD~\cite{kumagai2021semi,shi2022h2,zhang2022dual} because we do not have access to any annotated anomaly labels during the training phase. 


\section{Methodology}
\subsection{Framework Overview}

\begin{figure*}[ht]
    \centering
    \includegraphics[width=\textwidth]{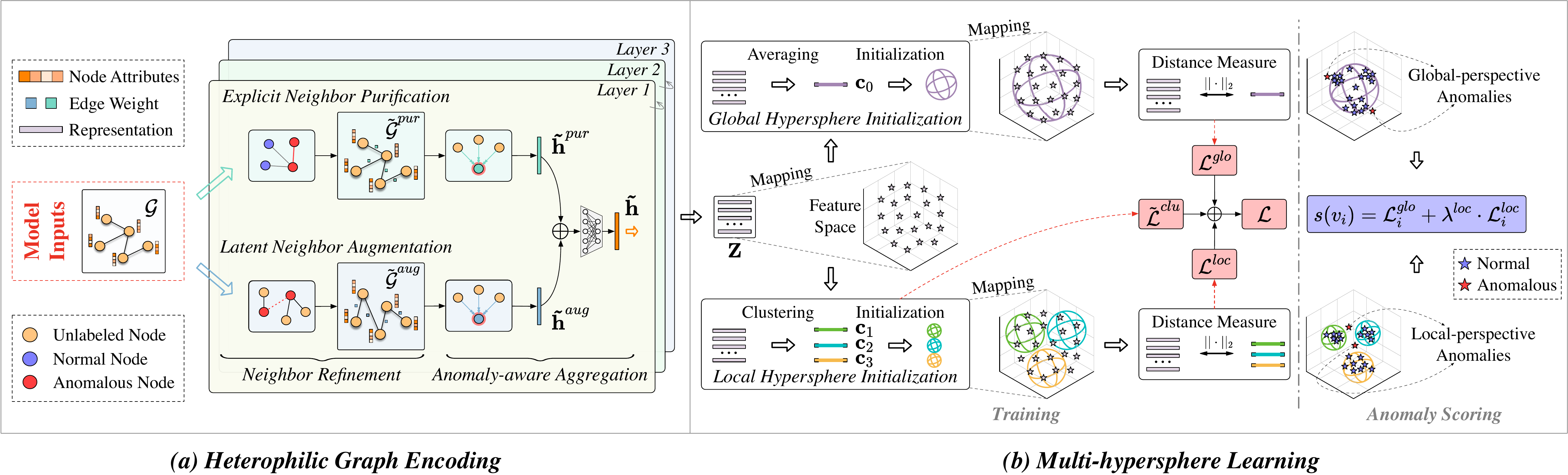}
    \caption{The overall architecture of MHetGL.}
    \label{figure:main}
\end{figure*}


\eat{The overall architecture of our proposed MHetGL is illustrated in Figure~\ref{figure:main}, which includes two stages: representation learning and anomaly scoring. (a) In the stage of representation learning, we propose \textsl{Heterophilic Graph Encoder} (HGE) to derive anomaly-aware representations in an unsupervised manner, which utilizes heterophilic graph convolutions via homophily-guided neighborhood refinement. (b) In the stage of anomaly identification, we propose \textsl{Multi-hypersphere learning} (MHL) to jointly consider local and global perspectives, which combines local and global hyperpshere learning for optimization and anomaly scoring. In the following sections, we will discuss the details of these modules and all the important notations are summarized in Appendix \ref{section:notation}.}

Figure~\ref{figure:main} illustrates the architecture of the proposed MHetGL framework. Overall, there are two major tasks in our approach: (1) learning node representations via a \textsl{Heterophilic Graph Encoding} (HGE) module, and (2) identifying anomalies through a \textsl{Multi-Hypersphere Learning} (MHL) module. In the first task, we develop HGE which consists of a homophily-guided neighborhood refinement block and an anomaly-aware aggregation block. Specifically, the neighborhood refinement block manipulates the graph structure by denoising the heterophilic edges and establishing latent connections between anomaly candidates. After that, the anomaly-aware aggregation block derives discriminative node representations by leveraging the refined graph structure for message passing and aggregation. In the second task, we introduce MHL, which devise multiple global and local hyperspheres for collective anomaly scoring. In addition, we also design a hypersphere regularization block to ensure the robust training of multiple hypersphere objectives. 
Totally, HGE and MHL modules correspond to two consecutive and complementary stages, and they are trained together in an end-to-end manner.
Next, we will discuss the detailed design of the above modules. 

\subsection{Heterophilic Graph Encoding}

\subsubsection{Homophily-guided neighborhood refinement}
We first elucidate the heterophily problem quantitatively. 
To measure the severity of heterophily for an attributed graph $\mathcal{G}=(\mathcal{V},\mathcal{E},\mathbf{X})$, we use the node-wise heterophily ratio $\mathcal{H}$ as the metric \cite{zheng2022graph}, which is formulated as:
\begin{equation}
    \mathcal{H}=\frac{1}{|\mathcal{V}|}\sum_{v\in\mathcal{V}}\frac{|\{u\in\mathcal{N}(v):y_u\neq y_v\}|}{|\mathcal{N}(v)|},
\end{equation}
where $y_v$ denotes the class label of node $v$. In particular, the heterophily issue of the anomalous nodes is much more severe than that of normals (\eg in Reddit, the average heterophily ratio of anomalous nodes is $81.53\%$, while that of the normal nodes is $0.55\%$). 

Based on such findings, we elaborate on how to refine the graph topology to facilitate learning distinguishable representations for graph anomalies. Concretely, our method is mainly motivated by two observations from the existing literature. First, given an anomalous node, the impact of heterophilic neighbors (\ie normal neighbors) should be reduced as they introduce noisy information to the node representation~\cite{shi2022h2}. Second, anomalous nodes that exhibit high structural and semantic similarities may be sparsely distributed across the graph and distant from each other. Taking such latent relationships into account is beneficial for GAD~\cite{liu2021pick}. Motivated by these observations, we propose two training-free graph topology refinement schemes: (1) purify explicit neighbors for anomalous nodes and (2) augment latent connections between anomalous nodes. 
Notably, the ground-truth anomalous nodes are unavailable under the unsupervised GAD setting. Therefore, during the neighborhood refinement process, we refine the neighbors of anomalous node candidates rather than exact anomalous nodes.
In this paper, we treat all the nodes as anomalous node candidates. Though this candidate selection strategy seems biased as most nodes in the graph are normal, it will not affect the quality of normal representations. As mentioned in the last paragraph, normal nodes usually exhibit strong homophily, thus their node representations can be further enhanced during the message passing process, demonstrating good distinguishability regardless of the neighborhood refinement block.

\begin{figure}[t]
\centering
\subfigure[Graph curvature distribution.]{\includegraphics[width=0.35\linewidth]{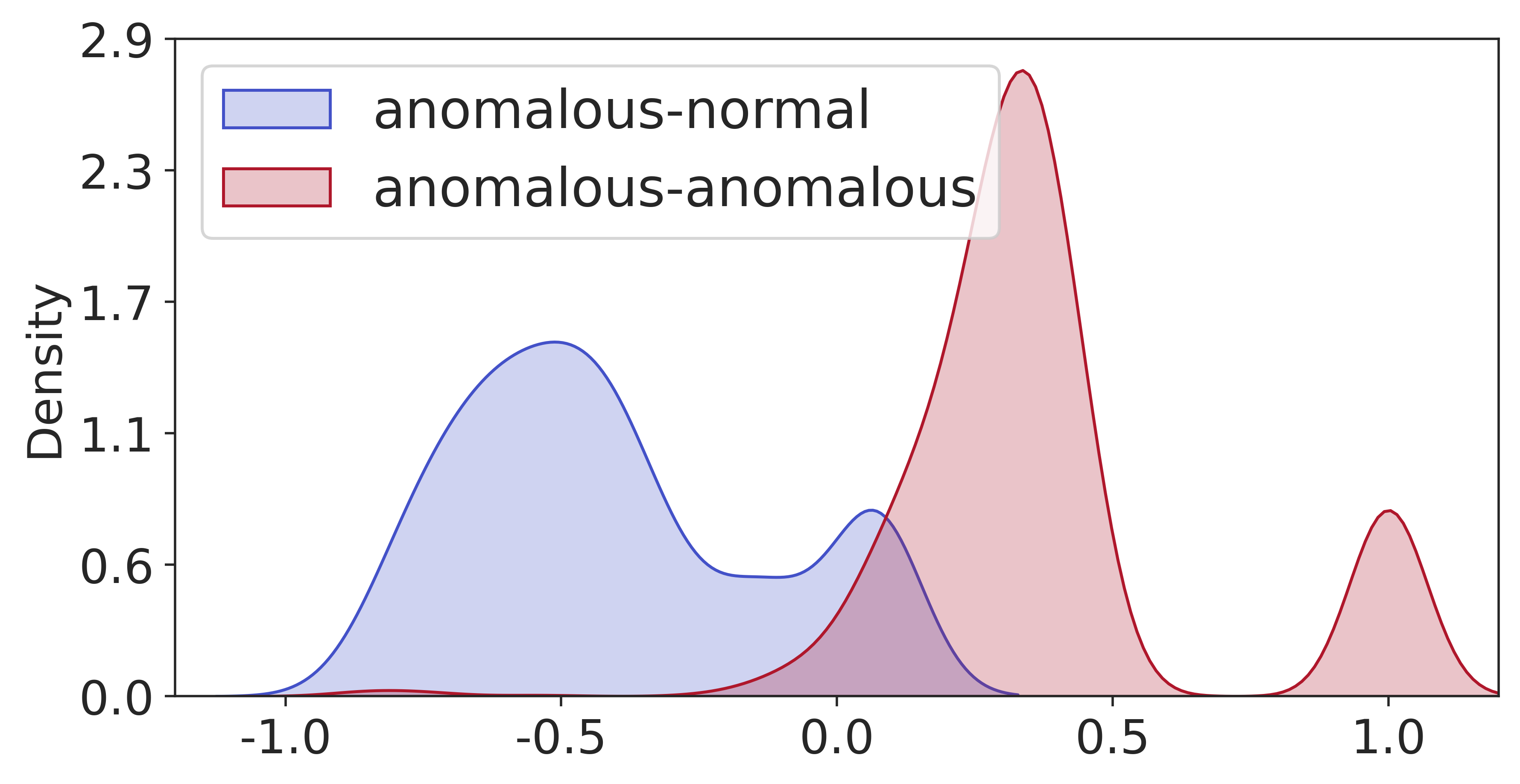}\label{figure:motivation_kde_2}}
\subfigure[GDV similarity distribution.]{\includegraphics[width=0.35\linewidth]{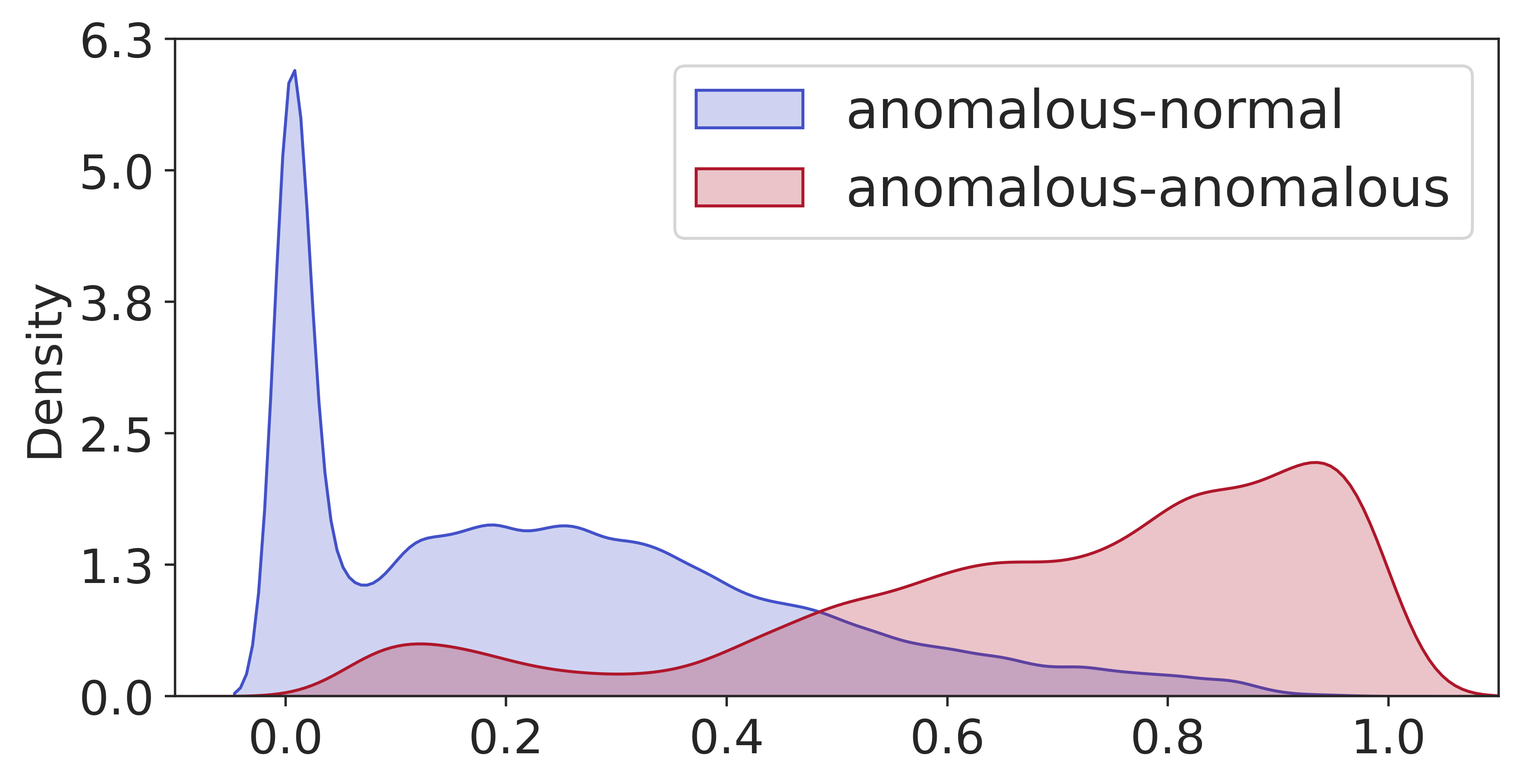}\label{figure:motivation_kde_3}}
\caption{The distribution discrepancy of two proposed measurements between anomalous-normal edges and anomalous-anomalous edges in the Citeseer dataset.} 
\label{figure:motivation_kde}
\end{figure}

\begin{figure}[h]
    \centering
    \subfigure[Graph curvature distribution.]{\includegraphics[width=0.35\linewidth]{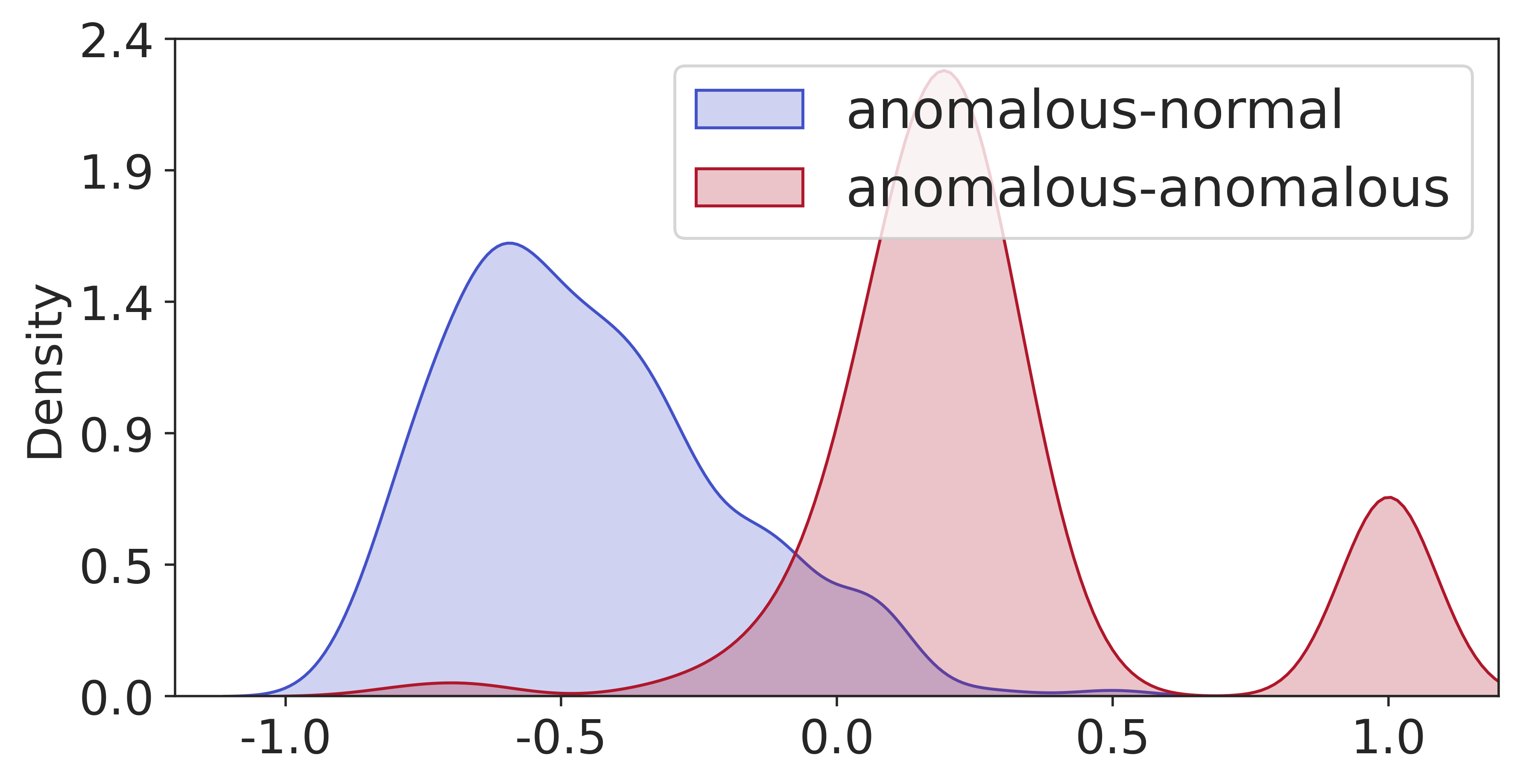}}
    \subfigure[GDV similarity distribution.]{\includegraphics[width=0.35\linewidth]{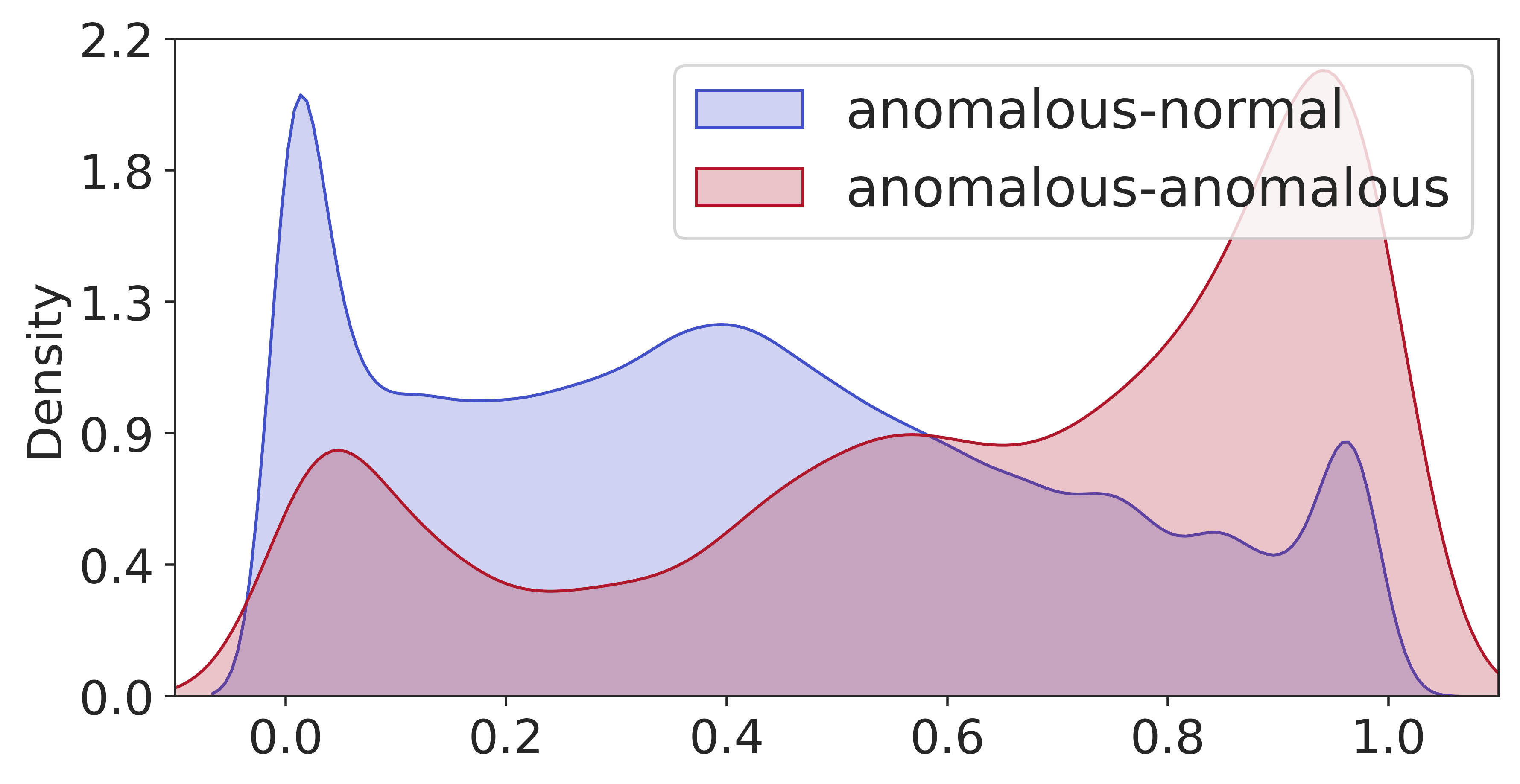}}
    \caption{The distribution of two proposed measurements in the Cora dataset.}
    \label{figure:motivation_cora}
\end{figure}

\begin{figure}[h]
    \centering
    \subfigure[Graph curvature distribution.]{\includegraphics[width=0.35\linewidth]{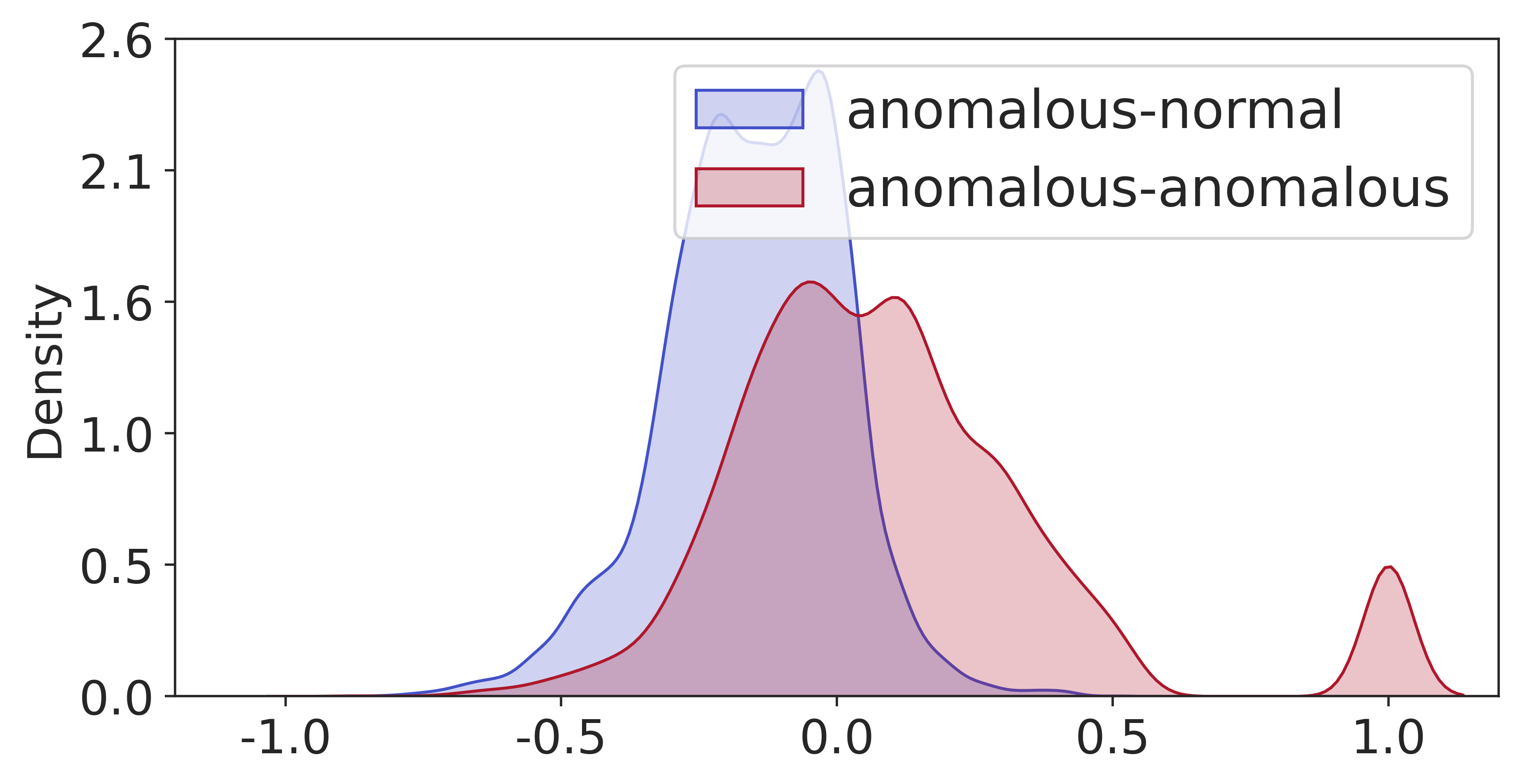}}
    \subfigure[GDV similarity distribution.]{\includegraphics[width=0.35\linewidth]{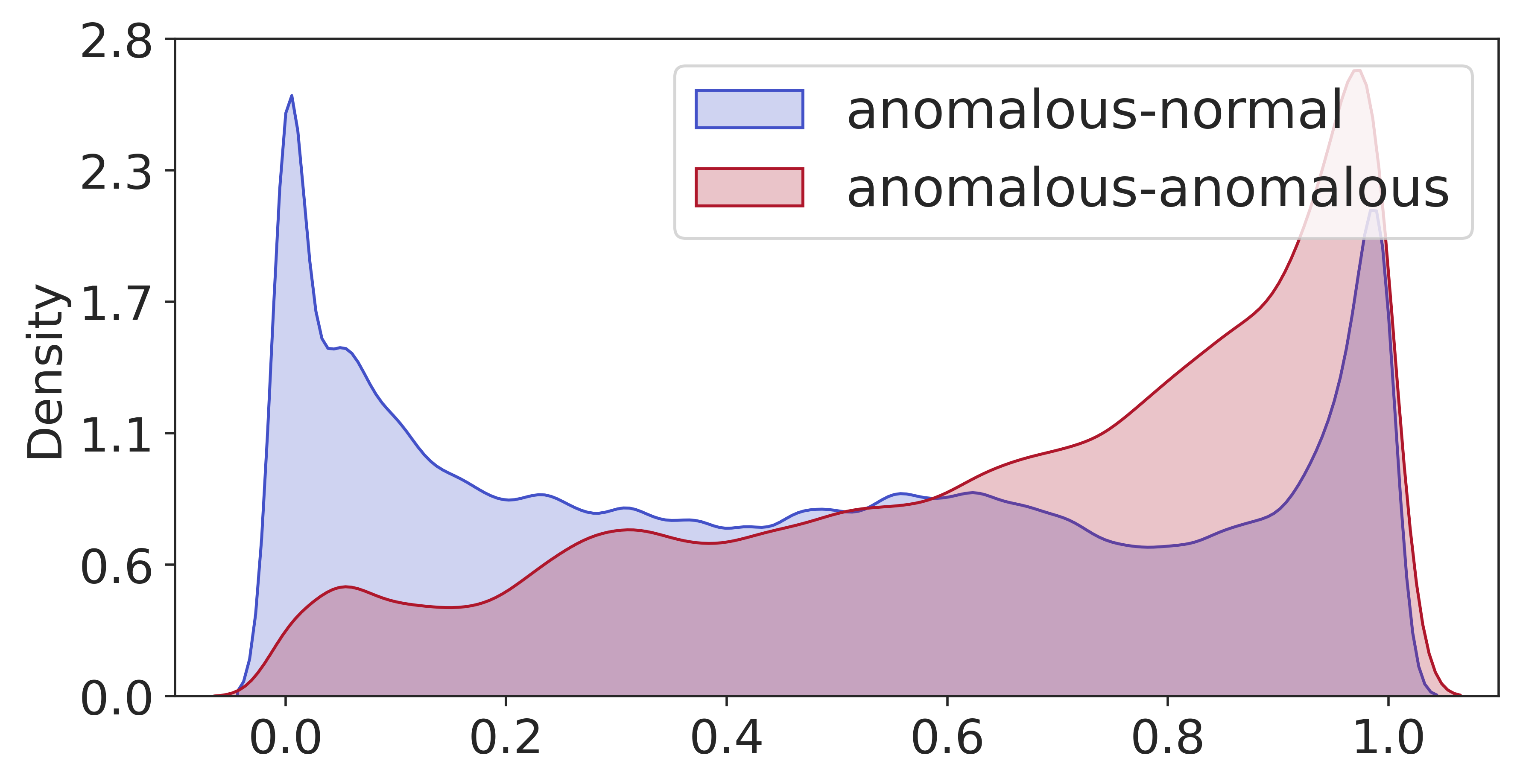}}
    \caption{The distribution of two types of edge weights in the Weibo dataset.}
    \label{figure:motivation_weibo}
\end{figure}

\noindent\textbf{Explicit neighborhood purification.} 
Intuitively, the homophily probabilities between the anomalous node and its neighbors (\ie the probabilities that the neighbors are also anomalous) are strongly correlated with the local structure characteristics. That is, the local structure around two connected anomalous nodes tends to be more complex~\cite{liu2022benchmarking}.
As validated in~\cite{chatterjee2021detecting}, graph curvature~\cite{ollivier2009ricci} can be utilized to measure the densely connected abnormal structural patterns among adjacent nodes.
Thus, we leverage graph curvature as a proxy for neighborhood purification under the unsupervised setting.
Specifically, we calculate the graph curvature $\kappa(i,j)$ for the edge between node $v_i$ and $v_j$ by
\begin{equation}
\begin{aligned}
&\mathbf{m}_i[o]=
\begin{cases}
\tau, & v_o=v_i, \\
(1-\tau)/|\mathcal{N}|, & v_o \in \mathcal{N}, \\
0, & otherwise,\\
\end{cases}
\\ &\kappa(i,j)=1-W(\mathbf{m}_i,\mathbf{m}_j),
\end{aligned}
\label{curv}
\end{equation}
where $o=1,2,...,N$ indicates the component index of the distribution vector $\mathbf{m}$, $\tau$ is a pre-defined coefficient within $[0,1]$, $W(\cdot,\cdot)$ is the Wasserstein distance, and $\mathcal{N}$ denotes the neighbors of $v_i$ in the original graph. 

To demonstrate the effectiveness of graph curvature based neighborhood purification, we visualize the curvature distribution of normal and anomalous nodes in the real-world Citeseer dataset by KDE plot, as depicted in Figure~\ref{figure:motivation_kde_2}.
As can be seen, the graph curvature distributions of anomalous and normal nodes fall in significantly different ranges. Specifically, the curvature of edges between anomalous nodes tend to be positive while the curvatures among anomalous-normal edges mainly fall in a negative range.
The visualization results of other datasets are similar (such as Figure~\ref{figure:motivation_cora} and Figure~\ref{figure:motivation_weibo}), which prove the observed patterns are universal knowledge.

\begin{figure}[h]
    \centering
    \subfigure[Converged weight distribution of normal and anomalous nodes in GREET~\cite{liu2023beyond}.]{\includegraphics[width=0.35\linewidth]{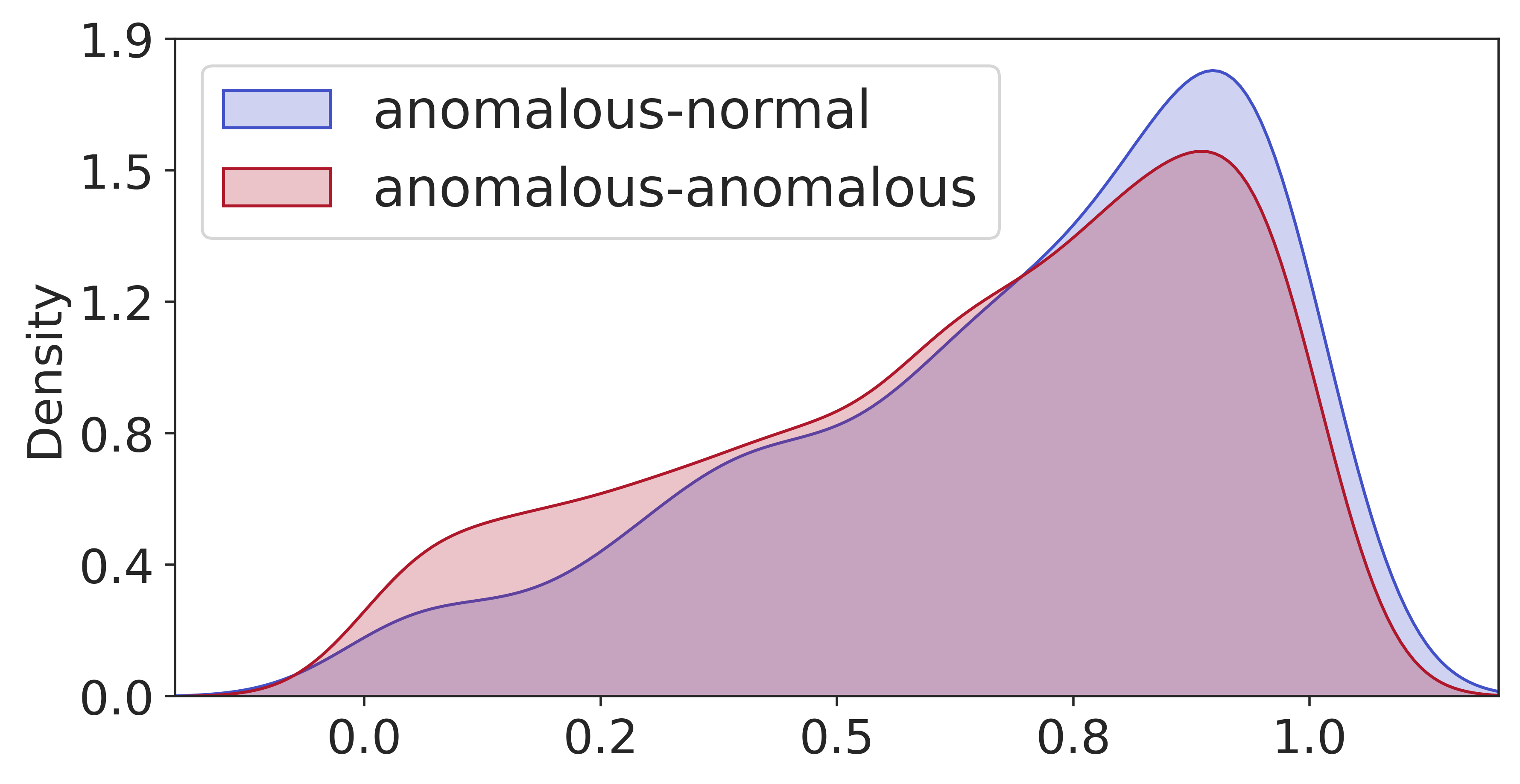}\label{figure:appendix_kde_1}}
    \subfigure[Diverged graph curvature distribution of normal and anomalous nodes.]{\includegraphics[width=0.35\linewidth]{fig/kdeplot/curvs_compare_an-aa_citeseer-syn_v1.png}\label{figure:appendix_kde_2}}
    \caption{The comparison of GREET and our neighborhood purification block.}
\end{figure}

Compared with general unsupervised heterophilic graph learning methods such as GREET~\cite{liu2023beyond}, graph curvature is more suitable for distinguishing anomalous-normal and anomalous-anomalous links for GAD tasks. 
In specific, GREET~\cite{liu2023beyond} proposes a learnable edge discriminator for unsupervised neighborhood purification in the graph with heterophily. We collect the low-pass edge weights of a GREET model trained in the real-world Citeseer dataset~\cite{yang2016revisiting}, and visualize the weight distribution of normal and anomalous nodes by kernel density estimate (KDE) plot, as illustrated in Figure~\ref{figure:appendix_kde_1}. Compared to our neighborhood purification block, we observe that GREET fails to distinguish normal and anomalous neighbors, because the class imbalance characteristic of graph anomalies is disregarded.
Such observations support our utilization of graph curvature as a tailored and effective proxy variable to distinguish heterophilic and homophilic neighbors of graph anomalies for neighbor purification.

Formally, given the original adjacency matrix $\mathbf{A}$, we refine the edge weights of $\mathbf{A}$ with graph curvature
\begin{equation}
\begin{aligned}
&\mathbf{A}^{pur}_{i,j}=
\begin{cases}
\kappa(i,j), & \mathbf{A}_{i,j}=1\\
0, & \mathbf{A}_{i,j}=0
\end{cases},\\
&\mathbf{\tilde{A}}^{pur}_{i,j}=\frac{\exp(\mathbf{A}^{pur}_{i,j})}{\sum_{j^\prime\in\tilde{\mathcal{N}}^{pur}_i}\exp(\mathbf{A}^{pur}_{i,j^\prime})},
\end{aligned}
\label{discriminate_weight}
\end{equation}
where $\tilde{\mathcal{N}}^{pur}_i=\tilde{\mathcal{N}}_i=\mathcal{N}_i\cup\{i\}$. By doing so, the edges connecting anomalies with heterophilic and homophilic neighbors are assigned with different weights, and the neighbors of anomalous nodes are purified, thus being more distinguishable. Since negative edge weights may lead to unstable model training and degrade the performance~\cite{li2022curvature}, here we apply a node-wise Softmax operation in Equation~\eqref{discriminate_weight} to normalize the weights and avoid generating the negative weights.
The impact of negative edge weights is analyzed in Section~\ref{section:neg_weight}.

\noindent\textbf{Latent neighbor augmentation.}
Furthermore, to reinforce the homophilic neighborhood aggregation effect, it is beneficial to augment the graph structure by connecting distant anomalous nodes.
Graphlet refers to a small induced subgraph that characterizes local topology~\cite{harshaw2016graphprints}, which can be leveraged to detect the occurrence of abnormal behaviors in the graph. 
In this work, we utilize the graphlet features to connect similar but distant anomalous nodes. Specifically, we calculate the occurrences of all surrounding graphlets consisting of up to $T$ nodes for each node $v_i$, and record the values via a Graphlet Degree Vector (GDV) $\mathbf{r}_i$~\cite{zhu2021community}, to indicate the structural role of the node. 

To illustrate the potential of GDV for anomalous node augmentation, we conduct an empirical analysis using the Citeseer dataset~\cite{yang2016revisiting}, as reported in Figure~\ref{figure:motivation_kde_3}. Specifically, we calculate the cosine similarity of GDVs among two arbitrary nodes, which measures the likelihood that there exists a potential connection. As can be seen, the GDV similarity of anomalous node pairs exhibits a right-shift distribution, while a left-shift distribution among anomalous-normal node pairs. In other words, anomalies tend to have higher GDV similarity with other anomalies within the graph. Thus, we leverage GDV similarity to augment latent homophilic neighbors for anomalous nodes.

Formally, we first construct a binary adjacency matrix using GDV similarity based on the original adjacency matrix $\mathbf{A}$. To reduce the computational overhead, we sparsify the matrix by only preserving edges with similarity higher than a threshold $\delta$. 
In addition, it is remarkable that the similarity of GDV features is insensitive to the size of substructures (\eg the GDV similarity of two isolated nodes is $1$, while the GDV similarity of two nodes with fully-connected ego-network is also $1$). 
However, nodes with more complicated substructures should be paid more attention as they usually encompass more abnormal structural information. 
Hence, we introduce degree-based edge re-weighting to discern node pairs with varying substructure sizes and obtain the weighted structure $\mathbf{A}^{aug}$. 
Then, we normalize the weights across the neighbor set and achieve the final adjacency matrix $\mathbf{\tilde{A}}^{aug}$, which is defined as
\begin{equation}
\begin{aligned}
&\mathbf{A}^{aug}_{i,j}=
\begin{cases}
\text{sim}(\mathbf{r}_i,\mathbf{r}_j)\cdot(|\mathcal{N}_i|+|\mathcal{N}_j|), & \text{sim}(\mathbf{r}_i,\mathbf{r}_j)\geq \delta,\\
0, & otherwise,
\end{cases}\\
&\mathbf{\tilde{A}}^{aug}_{i,j}=\frac{\mathbf{A}^{aug}_{i,j}}{\sum_{j^\prime\in\tilde{\mathcal{N}}^{aug}_i}\mathbf{A}^{aug}_{i,j^\prime}},
\end{aligned}
\label{extend_weight}
\end{equation}
where $\text{sim}(\cdot,\cdot)$ ranging from 0 to 1 is cosine similarity function and $\delta$ denotes the threshold for sparsification.
In our experiments, $\delta$ is defined by manual search and we practically set $\delta$ as 1 to balance performance and efficiency.
$\tilde{\mathcal{N}}^{aug}_i=\mathcal{N}^{aug}_i\cup\{i\}$ and $\mathcal{N}^{aug}_i$ denotes the neighbor set of node $v_i$ in the augmented graph constructed by GDV similarity.
Notably, we calculate $\mathbf{\tilde{A}}^{pur}$ and $\mathbf{\tilde{A}}^{aug}$ beforehand in the pre-processing stage, where these matrices are transformed into the sparsified form. This facilitates the memory and computational efficiency of our method.

\begin{figure}[h]
    \centering
    \subfigure[Graph curvature.]{\includegraphics[width=0.35\linewidth]{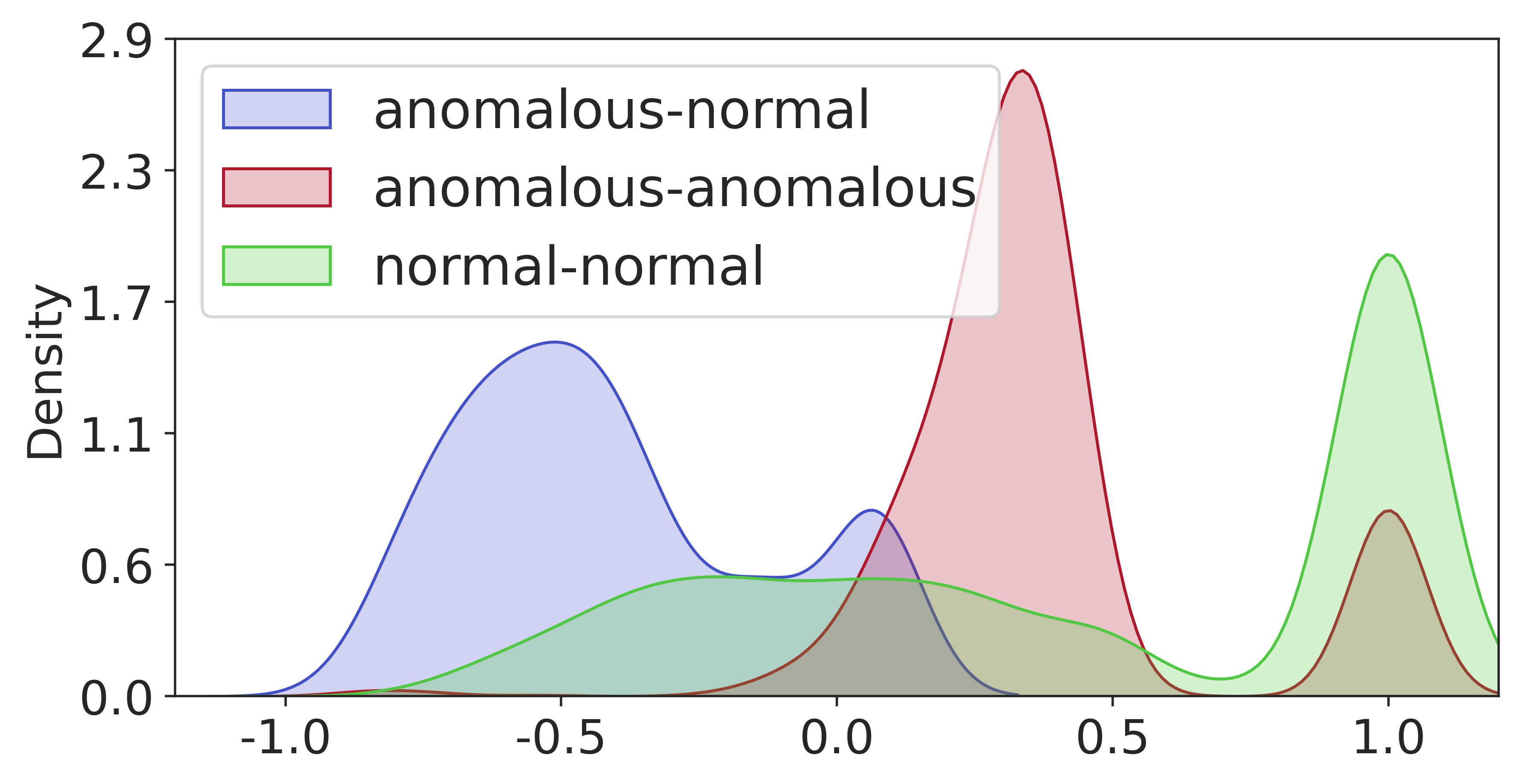}}
    \hspace{1pt}
    \subfigure[GDV similarity.]{\includegraphics[width=0.35\linewidth]{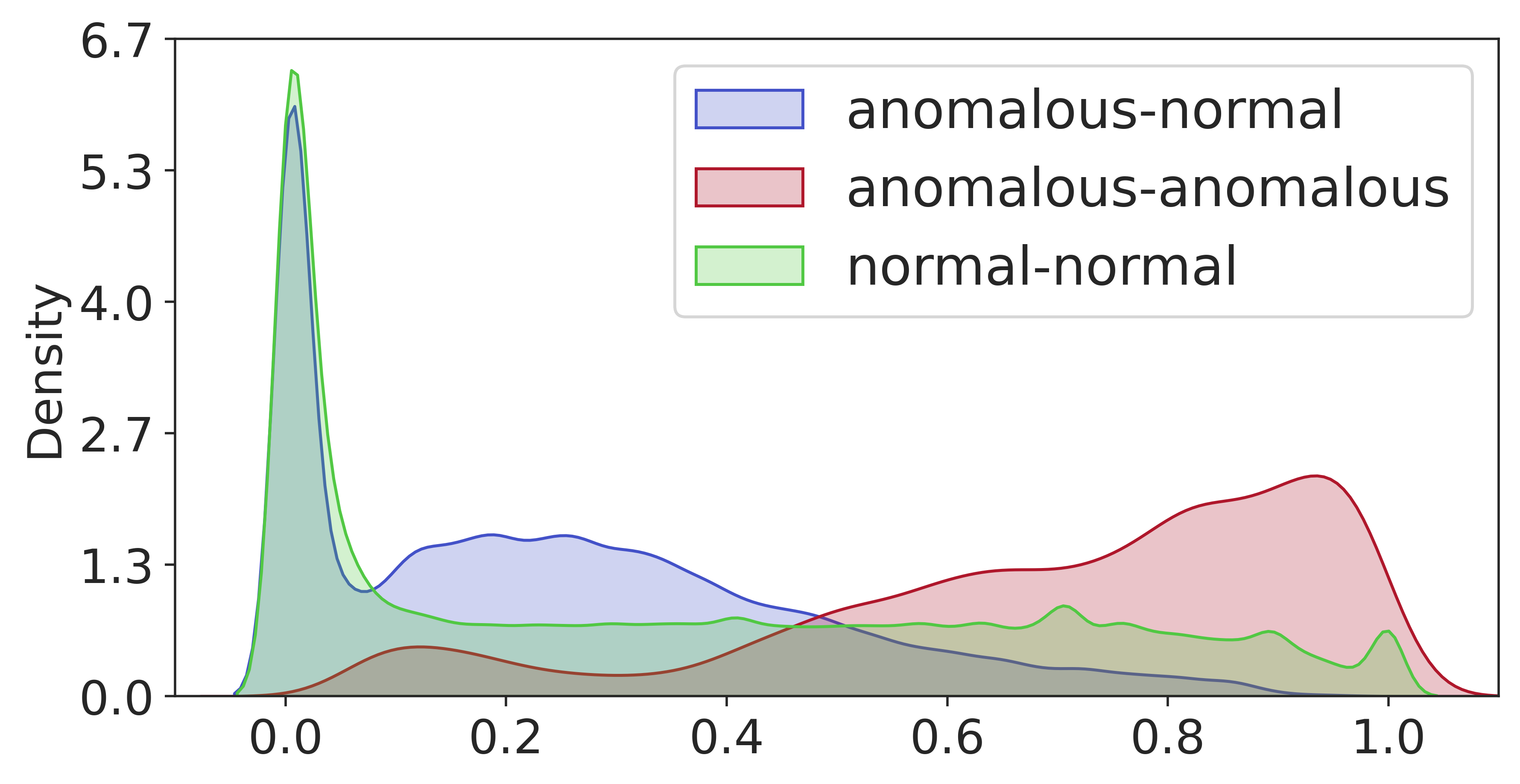}}
    \caption{The distribution of two types of edge weights in the CiteSeer dataset.}
    \label{figure:motivation_citeseer_complete}
\end{figure}

For clearness, we don't report the distribution of normal-normal edges as we focus on the property of anomalies in Figure~\ref{figure:motivation_kde}. Actually, the distribution of normal-normal edges is different from both abnormal-normal and abnormal-abnormal ones (Figure~\ref{figure:motivation_citeseer_complete}). In addition, we focus on the heterophily issue for anomalous nodes, as they are the primary cause of homophily-induced indistinguishability. As aforementioned, the distinguishability of normal nodes is unaffected by the proposed refinement block, thus we leave out the normal-normal edges.

\subsubsection{Anomaly-aware aggregation}
\eat{Based on the refined neighborhood represented by weighted graph structure $\mathbf{A}^{pur}$ and $\mathbf{A}^{aug}$, anomaly-aware aggregation is introduced for generating distinguishable representations. Specifically, we respectively apply anomaly-aware aggregation in $\mathbf{A}^{pur}$ and $\mathbf{A}^{aug}$, then fuse the two hidden outputs into the final representations.}
Building upon the refined graph structure, we propose an anomaly-aware aggregation module to generate distinguishable node embeddings. Specifically, we first conduct anomaly-aware aggregation on $\mathbf{\tilde{A}}^{pur}$ and $\mathbf{\tilde{A}}^{aug}$, then fuse the aggregation outputs to produce the final representations.

Here we take $\mathbf{\tilde{A}}^{pur}$ for illustration, the process on $\mathbf{\tilde{A}}^{aug}$ is similar. 
Specifically, we propose a two-fold aggregation operator to obtain representations~\cite{wijesinghe2021new} to exploit the anomaly-aware structural information and adaptively learn implicit neighbor correlation simultaneously. Concretely, the inputs of our aggregation module include the weighted adjacency matrix $\mathbf{\tilde{A}}^{pur}$ and the hidden representations $\mathbf{\tilde{h}}_i^{(l-1)}\in\mathbb{R}^d$ of node $v_i$ at layer $l-1$, where $d$ denotes the hidden dimension and $\mathbf{\tilde{h}}_i^{(0)}=\mathbf{x}_i\in\mathbb{R}^{d_{in}}$. To distinguish the anomalous nodes from normal ones, we aggregate the hidden representations with the anomaly-aware weights in the refined neighborhood
\begin{equation}
\begin{aligned}
{\mathbf{h}_i^{pur}}^{(l)}=\gamma^{(l)}\cdot \mathbf{\tilde{h}}_i^{(l-1)}+\sum_{j\in \mathcal{N}^{pur}_i} \frac{1+\mathbf{\tilde{A}}^{pur}_{i,j}}{\sqrt{|\tilde{\mathcal{N}}^{pur}_i||\tilde{\mathcal{N}}^{pur}_j|}}\cdot \mathbf{\tilde{h}}_j^{(l-1)},
\end{aligned}
\label{aggregation}
\end{equation}
where the coefficient $\mathbf{\gamma}^{(l)}\in\mathbb{R}$ is trainable. Then we further employ an attention layer to quantify the importance of different neighbors for aggregation, which is defined as 
\begin{equation}
\begin{aligned}
&\mathop{\mathbf{\tilde{h}}_i^{pur}}\nolimits^{(l)}=\sum_{j\in \tilde{\mathcal{N}}^{pur}_i}\alpha_{i,j}\cdot\mathbf{W}^{(l)}{\mathbf{h}_j^{pur}}^{(l)},\\
&\alpha_{i,j}=\frac{\exp(\text{ReLU}(\mathbf{a}^{(l)}\cdot[\mathbf{W}^{(l)}{\mathbf{h}_i^{pur}}^{(l)}||\mathbf{W}^{(l)}{\mathbf{h}_j^{pur}}^{(l)}]))}{\sum_{j^\prime\in \tilde{\mathcal{N}}^{pur}_i}\exp(\text{ReLU}(\mathbf{a}^{(l)}\cdot[\mathbf{W}^{(l)}{\mathbf{h}_i^{pur}}^{(l)}||\mathbf{W}^{(l)}{\mathbf{h}_{j^\prime}^{pur}}^{(l)}]))},
\end{aligned}
\label{gat}
\end{equation}
where $\mathbf{W}^{(l)},\mathbf{a}^{(l)}$ are trainable parameters and $\mathop{\mathbf{\tilde{h}}_i^{pur}}\nolimits^{(l)}\in\mathbb{R}^d$ denotes the hidden representation of node $v_i$ in the $l$-th layer.
Similarly, based on $\mathbf{A}^{aug}$ and the hidden representations $\mathbf{\tilde{h}}_i^{(l-1)}$. We apply the anomaly-aware aggregation and obtain the output $\mathop{\mathbf{\tilde{h}}_i^{aug}}\nolimits^{(l)}$.

Subsequently, we combine the node representations to derive the final representation to enhance the expressivity
\begin{equation}
\mathbf{\tilde{h}}_i^{(l)}=\mathbf{W}^{(l)}[\mathop{\mathbf{\tilde{h}}_i^{pur}}\nolimits^{(l)}||\ \mathop{\mathbf{\tilde{h}}_i^{aug}}\nolimits^{(l)}]+\mathbf{b}^{(l)},
\label{mlp}
\end{equation}
where $\mathbf{W}^{(l)},\mathbf{b}^{(l)}$ are trainable parameters, and $||$ denotes the concatenation operation. We define the output of the last layer as $\mathbf{z}_i=\mathbf{\tilde{h}}_i^{(L)}\in\mathbb{R}^d$, which will be utilized in the MHL module introduced below.





\subsection{Multi-hypersphere Learning}


In this section, we introduce the \textsl{Multi-Hypersphere Learning} (MHL) module.
We first describe hypersphere learning from a uniform global perspective and then extend it to multiple context-dependent local hypersphere learning.
Besides, we introduce the hypersphere regularization block to improve the robustness of the Multi-hypersphere Learning.

\subsubsection{Global hypersphere learning}
\eat{One-class classification \cite{wang2021one} has been extensively applied to GAD problem, which utilizes hypersphere learning to learn a compact hypersphere boundary to enclose all normal representations and detect anomalies outside the hypersphere. Specifically, we use the vanilla hypersphere objective to enclose all the normal node representations from the global perspective, which is defined as}

We first identify anomalies through advanced hypersphere learning. The key idea is to learn a compact hypersphere boundary to enclose all normal nodes in the latent space and identify the nodes outside the hypersphere as anomalies. In specific, we leverage the vanilla hypersphere objective to map all the normal node representations into a global hypersphere, formulated as
\begin{equation}
\begin{aligned}
\mathcal{L}^{glo}&=\frac{1}{|\mathcal{V}^{train}|} \sum_{v_i\in \mathcal{V}^{train}}\mathcal{L}^{glo}_i\\
&=\frac{1}{|\mathcal{V}^{train}|} \sum_{v_i\in \mathcal{V}^{train}} ||\mathbf{z}_i-\mathbf{c}_0||_2^2,
\end{aligned}
\label{HL_global}
\end{equation}
where $\mathbf{c}_0$ is the center of the global hypersphere. To derive the center, we examine three simple yet effective methods in Section~\ref{section:parameter}, demonstrating that our method is robust to the center initialization.

\subsubsection{Local hypersphere learning}
As aforementioned, anomalies may exist depending on local contexts in graphs (\eg local communities), which are overlooked by global hypersphere learning. To enable hypersphere learning with the ability to distinguish anomalies in the context of local graph structure, we propose first discovering the communities in the graph and then learning the compact hypersphere by considering the community structure. Particularly, there arise two key questions: (1) \textsl{how to detect communities over the graph without supervision signals}? (2) \textsl{how to integrate the community into hypersphere learning}?

Motivated by \cite{zhong2020deep}, we propose a contrastive-based clustering method for community detection, by maximizing the mutual information between the assignment distributions of nodes and their augmentations. Firstly, based on the node representation $\mathbf{z}_i$ derived from HGE, we learn the node-wise assignment probability vector
\begin{equation}  
\mathbf{p}_i=\text{Softmax}(\Phi_c(\mathbf{z}_i)),
\label{HL_cluster}
\end{equation}
where $\Phi_c(\cdot)$ is the assignment encoder and each component $\mathbf{p}_i[k]$ denotes the probability value that $v_i$ belongs to the cluster $C_k$. We assume there are $K$ clusters (\ie communities) and $\mathbf{p}_i\in \mathbb{R}^K$. Since community detection focuses on the local structure \cite{xu2007scan}, we instantiate $\Phi_c$ using GNN (\eg GAT), which is sufficient for capturing the structural information. We stack all node-wise assignment vectors and denote the assignment matrix as $\mathbf{P}\in \mathbb{R}^{N\times K}$. Each column of $\mathbf{P}$ can also be represented as $\mathbf{q}_k\in\mathbb{R}^N,k=1,2,\dots,K$, which is the cluster-wise assignment probability vector. To separate all the clusters, the assignment vectors of any two clusters, $\mathbf{q}_i$ and $\mathbf{q}_j$, are expected to be orthogonal to each other. This can be achieved by using contrastive learning~\cite{huang2020deep}. Thus, we directly contrast the cluster-wise assignment vectors with an augmentation-based objective
\begin{equation}
\mathcal{L}^{clu}=-\frac{1}{K}\sum_{k=1}^K \log \frac{\exp(\text{sim}(\mathbf{q}_k,\mathbf{q}_k^+))}{\sum_{k^{\prime}=1}^K\exp(\text{sim}(\mathbf{q}_k,\mathbf{q}_{k^{\prime}}^+))},
\label{HL_contrastive}
\end{equation}
where $\text{sim}(\cdot,\cdot)$ is the cosine similarity function and $\mathbf{q}^+$ is the column of the augmented assignment matrix $\mathbf{P}^+$. Note that augmentation plays a vital role in contrastive learning, and a plausible way for augmenting $\mathbf{P}$ is to perturb the graph structure (\eg dropping the nodes or edges)~\cite{you2020graph}. However, the typical augmentation strategy may break the semantics of graphs~\cite{lee2022augmentation}. Moreover, graph anomalies highly rely on the graph structure, which can be severely distorted by typical graph augmentation. As a result, rather than perturbing the graph structure, we augment the assignment distribution of nodes directly inspired by~\cite{bo2020structural}. Precisely, we augment each row of $\mathbf{P}$, \ie node-wise assignment vector as follows:
\begin{equation}
\begin{aligned}
&\mathbf{\tilde{p}}_i[k]=\frac{\mathbf{p}_i[k]^2}{f_k},k=1,2,\dots,K,\\
&\mathbf{p}_i^+=\text{softmax}(\mathbf{\tilde{p}}_i),
\end{aligned}
\label{HL_augment}
\end{equation}
where $f_k=\sum_{i=1}^N \mathbf{p}_i[k]$ are soft cluster frequencies.
Intuitively, node-wise assignment values in $\mathbf{P}$ are squared, thus further enhancing the gap between the assignments of high-confidence and low-confidence clusters. Soft cluster frequencies $f_k$ can help the augmented distribution assign higher values for low-frequency clusters to achieve a balance. The contrast between $\mathbf{P}$ and $\mathbf{P}^+$ can be regarded as a self-training mechanism, by which the representations within the same cluster will get closer. 


To incorporate the community perspective into hypersphere learning, we treat each community as a local hypersphere and initialize the local hypersphere centers with community representations. In particular, we compute the community representations via weighted summation~\cite{zong2018deep} as follows:
\begin{equation}
\mathbf{c}_k=\frac{\sum_{i=1}^N \mathbf{p}_i[k] \cdot \mathbf{z}_i}{\sum_{i=1}^N \mathbf{p}_i[k]}.
\label{HL_com_embed}
\end{equation}
Given the community representations, we can jointly constrain the node representations $\mathbf{z}_i,\ i=1,2,...,|\mathcal{V}|$ from various local community perspectives by optimizing the following objective
\begin{equation}
\begin{aligned}
\mathcal{L}^{loc}&=\frac{1}{|\mathcal{V}^{train}|} \sum_{v_i\in \mathcal{V}^{train}}\mathcal{L}^{loc}_i\\
&=\frac{1}{|\mathcal{V}^{train}|}\sum_{v_i\in\mathcal{V}^{train}}||\mathbf{z}_i-\mathbf{c}_{k_i^*}||_2^2,
\end{aligned}
\label{HL_local}
\end{equation}
where $k_i^*=\arg\max_k \mathbf{p}_i[k],k=1,2,\dots,K$.

\subsubsection{Hypersphere regularization} Note that hypersphere learning as described in Equations~\eqref{HL_global} and~\eqref{HL_local} is unstable and may encounter severe collapse problem without explicit constraints~\cite{qiu2022raising}. 
That is, the model objective, which aims to enclose the graph nodes within the hypersphere, can be achieved perfectly when the encoder maps all node representations to the center and the hypersphere's volume becomes zero (\ie $\Phi
_g(\mathbf{x}_i)=\mathbf{z}_i\equiv\mathbf{c},\ \forall v_i\in\mathcal{V}^{train}$, where $\Phi
_g(\cdot)$ denotes the graph encoder, $\mathbf{c}$ is the hypersphere center), which leads to a trivial solution and causes severe performance degradation. 

Some recent works, such as MSCL~\cite{reiss2023mean} and OCGTL~\cite{qiu2022raising}, have introduced auxiliary contrastive loss to address the collapse problem. However, they all focus on single hypersphere learning and have difficulty balancing diverse contrastive objectives across multiple hyperspheres. Moreover, OCGTL \cite{qiu2022raising} utilizes additional GNN modules for contrastive learning, which cannot be scaled to multi-hypersphere learning settings. To overcome the above limitations, we introduce a simple yet effective hypersphere regularization strategy. Concretely, since contrastive loss is utilized for community detection, we directly modify the clustering objective in Equation~\eqref{HL_contrastive} instead of supplementing a new regularization objective, which is formulated as follows:
\begin{equation}
\tilde{\mathcal{L}}^{clu}=-\frac{1}{K}\sum_{k=1}^K \log \frac{\exp(\text{sim}(\mathbf{c}_k,\mathbf{c}_k^+))}{\sum_{k^{\prime}=1}^K\exp(\text{sim}(\mathbf{c}_k,\mathbf{c}_{k^{\prime}}^+))},
\label{HL_contrastive_final}
\end{equation}
where $\mathbf{c}_k^+=(\sum_{i=1}^N \mathbf{p}_i^+[k]\cdot \mathbf{z}_i) / \sum_{i=1}^N \mathbf{p}_i^+[k]$.  

\begin{prop}
In global hypersphere learning, with the regularization of $\tilde{\mathcal{L}}^{clu}$, the constant graph encoder $\Phi_g(\mathbf{x}_i)\equiv\mathbf{c}_0,\ \forall v_i\in\mathcal{V}^{train}$ do not minimize $\mathcal{L}^{glo}$.
\label{proposition:global_collapse}
\end{prop}

\begin{prop}
In local hypersphere learning, with the regularization of $\tilde{\mathcal{L}}^{clu}$, the constant graph encoder $\Phi_g(\mathbf{x}_i)\equiv\mathbf{\tilde{c}},\ \forall v_i\in\mathcal{V}^{train}$ do not minimize $\mathcal{L}^{loc}$, for any $\mathbf{\tilde{c}}\in\{\mathbf{c}_1,...,\mathbf{c}_K\}$.
\label{proposition:local_collapse}
\end{prop}

Proposition~\ref{proposition:global_collapse} and \ref{proposition:local_collapse} show that Equation \eqref{HL_contrastive_final} can help avoid the collapse of both global and local hyperspheres. Intuitively, Equation \eqref{HL_contrastive_final} directly regularizes the representations, which ensures the directions of the representation vectors with respect to the hypersphere center are diversified by contrastive learning, and the representations will not collapse to the hypersphere center. The theoretical proofs are provided below.
Besides, instead of contrasting the cluster-wise assignment vectors, we directly contrast the community representations, which is highly scalable because it is independent of the number of hyperspheres.

\begin{proof}
(Proposition~\ref{proposition:global_collapse}).
We assume that global hypersphere collapse happens, which means the Equation~\eqref{HL_global} attains the minimum zero, \ie $\Phi_g(x_i)\equiv\mathbf{c}_0$. In this case, all the community representations $\mathbf{c}_k,k=1,2,...,K$ are the same, given the assignment encoder $\Phi_c$ is shared by all the nodes. Likewise, the augmented representations $\mathbf{c}_k^+,k=1,2,...,K$ are the same. We can assign a scalar $R$ for the fixed similarity $\text{sim}(\mathbf{c}_k,\mathbf{c}_k^+),k=1,2,...,K$, as such the clustering objective can be formulated as:
\begin{equation}
\begin{aligned}
\tilde{\mathcal{L}}^{clu}&=-\frac{1}{K}\sum_{k=1}^K \log \frac{\exp(\text{Sim}(\mathbf{c}_k,\mathbf{c}_k^+))}{\sum_{k^{\prime}=1}^K\exp(\text{Sim}(\mathbf{c}_k,\mathbf{c}_{k^{\prime}}^+))}\\
&=\log K.
\end{aligned}
\end{equation}
The global hypersphere objective $\mathcal{L}^{glo}$ aims at reducing the distances between the representations and the center $\mathbf{c}_0$. Since the community representation is the weighted sum of node representations, each $\mathbf{c}_k$ is encouraged to approach the center of the global hypersphere. If we fix the distances of community representations respect to the center, the contrastive loss $\tilde{\mathcal{L}}^{clu}$ can also be minimized, given that cosine similarity $\text{Sim}(\mathbf{c}_k,\mathbf{c}_k^+)$ only captures the angular information and be insensitive to the distances. Therefore, as long as the loss $\tilde{\mathcal{L}}^{clu}\textless\log K$, it will not be disturbed by the global objective and rather keep decreasing during training, thus the collapse does not happen. The assumption $\tilde{\mathcal{L}}^{clu}\textless\log K$ can be verified in practical experiments. 
\label{proof:1}
\end{proof}

\begin{proof}
(Proposition~\ref{proposition:local_collapse}).
We have many local hyperspheres $C_1,...,C_K$, with center representations $\mathbf{c}_1,...,\mathbf{c}_K$. Without loss of generality, we assume hypersphere collapse happens in $C_1$ and the nodes belonging to community $C_1$ constitute the set $S_1=\{v_i|\arg\max_k \mathbf{p}_i[k]=1,k=1,2,...,K\}$ with cardinality $\textgreater 1$. To satisfy $\Phi_g(\mathbf{x}_i)\equiv\mathbf{c}_1,i\in S_1$, the graph feature encoder $\Phi_g$ must be constant. So for all the nodes $v_i\in \mathcal{V}^{train}$, we have $\Phi_g(\mathbf{x}_i)\equiv\mathbf{c}_1$ and then all the communities coincide. By minimizing the loss $\mathcal{L}^{glo}$, all the representations collapse to global hypersphere center. Therefore the local hypersphere collapse is equivalent to global hypersphere collapse, and could be prevented showed in Proof~\ref{proof:1}.
\end{proof}

\subsection{Training and Anomaly Scoring} 
In MHetGL, we aim to jointly minimize the following objectives:
\begin{equation}
\mathcal{L}=\mathcal{L}^{glo}+\lambda^{loc}\cdot \mathcal{L}^{loc}+\lambda^{clu}\cdot \tilde{\mathcal{L}}^{clu},
\label{HL_total_loss}
\end{equation}
where $\lambda^{loc}$ and $\lambda^{clu}$ are pre-defined hyper-parameters that control the importance of the local hypersphere loss $\mathcal{L}^{loc}$ and the cluster loss $\tilde{\mathcal{L}}^{clu}$.

Finally, we measure the abnormality of each node from both local and global perspectives. In particular, we define the anomaly score of node $v_i$ as
\begin{equation}
s(v_i)=\mathcal{L}^{glo}_i+\lambda^{loc}\cdot \mathcal{L}^{loc}_i=||\mathbf{z}_i-\mathbf{c}_0||_2^2+\lambda^{loc}\cdot||\mathbf{z}_i-\mathbf{c}_{k_i^*}||_2^2.
\label{HL_score}
\end{equation}

\subsection{Computational Analysis}
This section presents the complexity analysis of our method.
In HGE, as we derive the sparsified weighted adjacency matrices $\mathbf{\tilde{A}}^{pur}$ and $\mathbf{\tilde{A}}^{aug}$ beforehand in the preprocessing step, the complexity of graph convolution for $\mathbf{\hat{A}}^{pur}$ and $\mathbf{\hat{A}}^{aug}$ are $O(Nd^2+|\mathcal{E}^{pur}|d)$ and $O(Nd^2+|\mathcal{E}^{aug}|d)$ respectively. Adding a linear network with complexity $O(Nd^2)$, the complexity of HGE layer becomes $O(Nd^2+|\mathcal{E}^{pur}|d+|\mathcal{E}^{aug}|d)$. In each layer, we can further parallize two convolution module with complexity $O(Nd^2+\max(|\mathcal{E}^{pur}|,|\mathcal{E}^{aug}|)d)$. For the first layer, the complexity is formulated as $O(Nd^{in}d+\max(|\mathcal{E}^{pur}|,|\mathcal{E}^{aug}|)d)$. 
In MHL, with the hypersphere regularization, the complexity of contrastive clustering objective can be reduced from $O(N)$ to $O(d)$. 
Overall, the complexity of our method is $O(N+\max(|\mathcal{E}^{pur}|,|\mathcal{E}^{aug}|))$, where we omit hidden dimension $d$ and number of clusters $K$.
\section{Experiments}
This section presents the experimental results of MHetGL, which includes the experimental setup, overall performance comparison, ablation study, parameter analysis, and case study. 

\subsection{Experimental Setup}\label{section:experimental setup}
\noindent\textbf{Datasets.} For node-level GAD, there exist two kinds of datasets: \textsl{Injected} and \textsl{organic} datasets \cite{liu2022benchmarking}. Since labeling high-quality ground-truth anomalies require intensive manual efforts, most of the existing works inject clean datasets and get injected data for evaluation. However, in this paper, we also consider six organic GAD datasets collected in the real world.
Specifically, we evaluate our model on nine widely used real-world graph datasets, including four injected datasets: Cora~\cite{yang2016revisiting}, Citeseer~\cite{yang2016revisiting}, ML~\cite{bojchevski2017deep}, Pubmed~\cite{yang2016revisiting}, and six organic datasets: Reddit~\cite{liu2022benchmarking}, Weibo~\cite{liu2022benchmarking}, Books~\cite{liu2022benchmarking}, Disney~\cite{liu2022benchmarking}, Enron~\cite{liu2022benchmarking}, Questions~\cite{tang2023gadbench}. For the first four datasets, we follow~\cite {liu2022benchmarking} and inject two kinds of injected anomalies: structural and contextual, with the ratio $1:1$. For the structural anomaly injection, we randomly select some non-overlapping groups of nodes as anomalies and make the nodes in each group fully-connected. 
For the contextual injection, we randomly select some nodes as anomalies and replace the attributes of each node with dissimilar attributes from another node in the same graph. 
Besides, the six latter datasets were originally equipped with ground-truth abnormal labels. We show the statistical information about all the datasets in Table \ref{table:data}.
We follow the strategy in OCGNN \cite{wang2021one} and split the dataset into training, validation, and test set by the ratio of $6:1:3$. 

\begin{table}[ht]
\centering
\caption{Statistical information of datasets.}
\resizebox{0.5\linewidth}{!}{
\begin{tabular}{c|ccccc}
\toprule
\textbf{Dataset} & \textbf{\#Nodes} & \textbf{\#Edges} & \textbf{\#Feat.} & \textbf{\#Ano.} & \textbf{Ano. Ratio}\\
\hline \hline
\textbf{Cora}
 & 2708 & 14844 & 1433 & 150 & 5.54\%\\
\hline
\textbf{CiteSeer}
 & 3327 & 14457 & 3703 & 169 & 5.08\%\\
\hline
\textbf{ML}
 & 2995 & 20893 & 2879 & 150 & 5.01\%\\
\hline
\textbf{PubMed}
 & 19717 & 137191 & 500 & 985 & 5.00\%\\
\hline
\textbf{Reddit}
 & 10984 & 168016 & 64 & 366 & 3.33\%\\
\hline
\textbf{Weibo}
& 8405 & 762947 & 400 & 868 & 10.33\%\\
\hline
\textbf{Books}
& 1418 & 8808 & 21 & 28 & 1.97\%\\
\hline
\textbf{Disney}
& 124 & 794 & 28 & 6 & 4.84\%\\
\hline
\textbf{Enron}
& 13533 & 367507 & 18 & 5 & 0.04\%\\
\hline
\textbf{Questions}
& 48921 & 356001 & 301 & 1460 & 2.98\%\\
\bottomrule
\end{tabular}
}
\label{table:data}
\end{table}

\noindent\textbf{Baselines.} We compare our proposed framework with the following baselines: a heuristic method DEG which directly uses node degree as the anomaly score, two traditional approaches including a density-based method LOF~\cite{breunig2000lof} and a clustering-based method SCAN~\cite{xu2007scan}, a residual-based approach Radar~\cite{li2017radar}, two contrastive learning-based approaches including CoLA~\cite{liu2021anomaly} and GRADAT~\cite{duan2023graph}, four state-of-the-art hypersphere learning-based approaches including OCGNN \cite{wang2021one}, AAGNN~\cite{zhou2021subtractive}, MHGL \cite{zhou2022unseen} and OCGTL~\cite{qiu2022raising}, four autoencoder-based approaches including Dominant~\cite{ding2019deep}, ComGA~\cite{luo2022comga}, GAD-NR~\cite{roy2023gad} and VGOD~\cite{huang2023unsupervised} and a novel detection method TAM~\cite{tam} based on local node affinity. In addition, GREET~\cite{liu2023beyond} which discriminates heterophilic edges is also considered, and we pass the representations learned by GREET model to our MHL to report the results.

\noindent\textbf{Metrics.} We use AUPR (\ie the area under precision-recall curve) and AUROC (\ie the area under ROC curve), two commonly used metrics for GAD evaluation~\cite{liu2022benchmarking,zhou2021subtractive,zhou2022unseen}. Notably, AUPR can adjust for samples with severe class imbalance issue and focuses on positive samples (\ie anomalous) compared to AUROC \cite{zhou2023improving}.

\begin{table}[ht]
\centering
\caption{Hyperparameter setup of datasets.}
\resizebox{0.35\linewidth}{!}{
\begin{tabular}{c|ccccc}
\toprule
\textbf{Dataset} & $\lambda^l$ & $\lambda^c$ & $K$ & $\mathbf{c}_0$ & $d$\\
\hline \hline
\textbf{Cora}
 & 1 & 0.1 & 8 & \textsl{Init.} & 32\\
\hline
\textbf{CiteSeer}
 & 0.001 & 10 & 6 & \textsl{Update} & 32\\
\hline
\textbf{ML}
 & 1 & 10 & 10 & \textsl{Init.} & 256\\
\hline
\textbf{PubMed}
 & 0.01 & 0.01 & 4 & \textsl{Update} & 32\\
\hline
\textbf{Reddit}
 & 1 & 10 & 8 & \textsl{Train} & 32\\
 \hline
\textbf{Weibo}
 & 0.001 & 0.01 & 8 & \textsl{Update} & 128\\
 \hline
\textbf{Books}
 & 1 & 1 & 6 & \textsl{Train} & 64\\
 \hline
\textbf{Disney}
 & 10 & 10 & 8 & \textsl{Init.} & 32\\
 \hline
\textbf{Enron}
 & 1 & 10 & 8 & \textsl{Train} & 256\\
 \hline
 \textbf{Questions}
 & 1 & 10 & 6 & \textsl{Update} & 256\\
\bottomrule
\end{tabular}
}
\label{table:hyperparam}
\end{table}
\noindent\textbf{Implementation Details}
\label{section:setup}
To facilitate reproducibility, we elaborate on the implementation details in this section. 
The graph attention layer is implemented by the toolkit PyG and we use LeakyReLU as the activation function in our model. We optimize our model with Adam and set the weight decay coefficient as 0.0005. Following OCGNN, we trained our model using an early stopping strategy on AUC score on the validation set, with a maximum of 10000 epochs and a patience of 1000 epochs. We set the number of layers of HGE as 2 and set 1 layer for the assignment encoder $\Phi_c$ in clustering. Moreover, we tune several significant parameters including loss weights $\lambda^l,\lambda^c$, number of clusters $K$, the setting of global hypersphere center and hidden size $d$, which are analyzed in Section \ref{section:parameter}. In Table \ref{table:hyperparam}, we show the experimental setup of hyperparameters on all the datasets.
\begin{table*}[ht]
\centering
\caption{Anomaly detection results (\%). OOM\_C(G) denotes out of the C(G)PU memory.}
\resizebox{\linewidth}{!}{
\begin{tabular}{c|c||c|c|c|c||c|c|c|c|c|c}
\toprule
\textbf{Method} & \textbf{Metric} & \textbf{Cora} & \textbf{CiteSeer} & \textbf{ML} & \textbf{PubMed} & \textbf{Reddit} & \textbf{Weibo} & \textbf{Books} & \textbf{Disney} & \textbf{Enron} & \textbf{Questions}\\
\hline \hline

\multirow{2}{*}{\textbf{DEG}}
& \textbf{AUPR} & 64.57 & 70.58 & 34.14 & 64.18 
& 3.72 & 5.90 & 1.73 & 6.62 & 0.05 & \underline{\textsl{5.34}}\\
& \textbf{AUROC} & 96.62 & 99.00 & 94.60 & 97.82
& 56.36 & 20.47 & 38.62 & 50.71 & 37.81 & \underline{\textsl{62.47}}\\
\hline

\multirow{2}{*}{\textbf{LOF}} 
& \textbf{AUPR} & 24.28 & 26.96 & 18.44 & 4.81
& 3.51 & 0.06 & 2.18 & 4.78 & 0.06 & 3.38\\
& \textbf{AUROC} & 69.58 & 68.67 & 46.58 & 20.98
& 51.77 & 42.11 & 48.41 & 17.14 & 42.11& 54.45\\
\hline

\multirow{2}{*}{\textbf{SCAN}} 
& \textbf{AUPR} & 5.31 & 4.92 & 5.93 & 8.25
& 3.28 & 16.65 & 2.30 & 5.56 & 0.05 & 2.85\\
& \textbf{AUROC} & 43.02 & 37.17 & 59.97 & 72.17
& 50.49 & 71.54 & 54.08 & 51.43 & 33.79 & 49.13\\
\hline

\multirow{2}{*}{\textbf{Radar}} 
& \textbf{AUPR} & 21.47 & 10.20 & 57.72 & 25.64
& 3.47 & 43.73 & 1.87 & 19.44 & 0.09 & OOM\_C\\
& \textbf{AUROC} & 76.58 & 71.43 & 97.99 & 76.04
& 50.78 & 45.44 & 39.09 & 48.57 & 59.80 & OOM\_C\\
\hline

\multirow{2}{*}{\textbf{CoLA}} 
& \textbf{AUPR} & 14.62 & 17.04 & 4.32 & 13.57
& \underline{\textsl{4.76}} & 8.01 & 2.11 & 19.52 & 0.05 & 3.18\\
& \textbf{AUROC} & 60.62 & 74.04 & 48.18 & 76.80
& \underline{\textsl{58.90}} & 39.03 & 50.00 & \underline{\textsl{72.14}} & 32.82 & 52.54\\
\hline

\multirow{2}{*}{\textbf{GRADATE}} 
& \textbf{AUPR} & 24.97 & 53.04 & 19.42 & OOM\_C
& 4.11 & 13.09 & 2.53 & 4.34 & 0.05 & OOM\_C\\
& \textbf{AUROC} & 73.27 & 88.59 & 62.75 & OOM\_C
& 59.12 & 38.55 & 46.14 & 7.14 & 32.49 & OOM\_C\\
\hline

\multirow{2}{*}{\textbf{OCGNN}}
& \textbf{AUPR} & 42.55 & 10.88 & 73.15 & 51.55
& 3.79 & 73.48 & 2.56 & 9.42 & 0.10 & 3.78\\
& \textbf{AUROC} & 89.00 & 76.82 & 98.66 & 96.79
& 58.23 & 90.96 & 47.81 & 53.57 & 45.08 & 59.50\\
\hline

\multirow{2}{*}{\textbf{AAGNN}}
& \textbf{AUPR} & 17.66 & 11.15 & 21.47 & 82.35
& 3.06 & 47.52 & 2.41 & 8.56 & 0.07 & 4.78\\
& \textbf{AUROC} & 83.81 & 80.31 & 86.45 & 95.91
& 49.20 & 83.12 & 50.71 & 57.00 & 46.28 & 56.69\\
\hline

\multirow{2}{*}{\textbf{MHGL}} 
& \textbf{AUPR} & 50.92 & 69.19 & 53.55 & 52.86
& 3.99 & 56.35 & 2.40 & 11.94 & 0.10 & \textbf{6.12}\\
& \textbf{AUROC} & 90.38 & 93.95 & 97.33 & 92.78
& 58.65 & 84.82 & 39.36 & 65.29 & 60.29 & 61.27\\
\hline

\multirow{2}{*}{\textbf{OCGTL}} 
& \textbf{AUPR} & 24.71 & 30.29 & 50.88 & 47.95
& 4.01 & \textbf{84.49} & 2.37 & 7.42 & 0.08 & 4.01\\
& \textbf{AUROC} & 89.42 & 87.12 & 95.84 & 95.68
& 58.76 & \underline{\textsl{97.26}} & 41.76 & 45.14 & 51.28 & 60.73\\
\hline

\multirow{2}{*}{\textbf{Dominant}} 
& \textbf{AUPR} & 57.25 & 62.34 & 33.29 & 60.41
& 3.61 & 75.05 & 2.27 & \underline{\textsl{19.79}} & 0.05 & OOM\_G\\
& \textbf{AUROC} & 96.16 & 98.71 & 95.00 & 98.27
& 56.59 & 90.52 & 50.20 & 54.29 & 29.50 & OOM\_G\\
\hline

\multirow{2}{*}{\textbf{ComGA}} 
& \textbf{AUPR} & 70.01 & 75.70 & 40.20 & 68.15
& 3.34 & 80.94 & 2.85 & 9.03 & \underline{\textsl{0.22}} & OOM\_G\\
& \textbf{AUROC} & 97.27 & 99.34 & 95.64 & 98.34
& 51.37 & 92.91 & 57.81 & 41.43 & 55.61 & OOM\_G\\
\hline

\multirow{2}{*}{\textbf{GAD-NR}} 
& \textbf{AUPR} & 54.68 & 65.52 & 29.89 & 7.67
& 3.28 & 7.83 & 1.72 & 7.54 & \textbf{0.31} & OOM\_G\\
& \textbf{AUROC} & 96.33 & 98.90 & 94.16 & 67.69
& 51.32 & 44.33 & 35.01 & 48.57 & \underline{\textsl{61.69}} & OOM\_G\\
\hline

\multirow{2}{*}{\textbf{VGOD}} 
& \textbf{AUPR} & \underline{\textsl{94.67}} & \underline{\textsl{85.26}} & \underline{\textsl{78.33}} & \textbf{98.45}
& 4.10 & 39.29 & \underline{\textsl{3.83}} & 4.25 & 0.05 & 2.99\\
& \textbf{AUROC} & \textbf{99.74} & \underline{\textsl{99.46}} & \underline{\textsl{98.90}} & \textbf{99.94}
& 52.53 & 58.01 & \underline{\textsl{60.62}} & 4.29 & 27.61 & 52.71\\
\hline \hline

\multirow{2}{*}{\textbf{MHetGL}} 
& \textbf{AUPR} & \textbf{96.67} & \textbf{99.22} & \textbf{90.74} & \underline{\textsl{92.87}}
& \textbf{5.09} & \underline{\textsl{81.62}} & \textbf{8.35} & \textbf{33.33} & 0.16 & 5.26\\
& \textbf{AUROC} & \underline{\textsl{99.51}} & \textbf{99.97} & \textbf{99.69} & \underline{\textsl{99.15}}
& \textbf{65.37} & \textbf{97.50} & \textbf{77.06} & \textbf{85.00} & \textbf{76.93} & \textbf{63.03}\\

\bottomrule
\end{tabular}
}
\label{table:main}
\end{table*}

\subsection{Overall Performance Comparison}

Table \ref{table:main} reports the overall performance of our method and all the
compared baselines on nine datasets with respect to two evaluation metrics. Overall, The proposed MHetGL significantly outperforms all the baselines on both the injected and organic datasets, which demonstrates its superiority for unsupervised GAD. 

For different types of baseline models, we can make the following observations. (1) Deep methods are generally better than non-deep and non-graph methods (\ie LOF, SCAN, and Radar), which demonstrates the significance of structural information and the superiority of deep learning in detecting graph anomalies. (2) The heuristic method DEG surprisingly outperforms most of the baselines on injected data with only node degree features, which verifies the finding of \cite{huang2023unsupervised} that there exists a serious data leakage issue of anomaly injection strategy~\cite{liu2022benchmarking}. How to conquer the leakage issue is orthogonal to our study which is left for future work. Generally, our method can outperform the heuristic method DEG with a significant improvement. (3) Autoencoder-based methods generally outperform hypersphere-based methods on injected data while behaving worse on organic data. It demonstrates that the reconstruction scheme is ready to detect feature-deviating or densely connected injected anomalies and is difficult to generalize to real-world anomalies, while hypersphere-based methods show superiority for real-world anomaly detection. Our method designs an advanced hypersphere learning module that shows superiority on all kinds of data.

For baseline methods, AUPR values can be low in specific data and settings, though AUROC values are high. A possible reason is that AUROC aims to evaluate the performance on positive (\ie anomalous) and negative (\ie normal) samples in a balanced way, and AUPR merely considers the positive samples. Thus, AUPR is more sensitive to the performance of anomaly identification and AUROC may overestimate the results. Compared to baselines, MHetGL can achieve both high AUROC values and high AUPR values on all the datasets. 

In addition, compared to injected data, it is hard to achieve decent results on organic data (\eg Reddit, Books, Enron). Also, the performance improvements observed on organic datasets are much more significant than those on injected datasets. This implies that real-world anomalies are much harder to identify than injected anomalies, and existing methods have already achieved satisfactory results on injected datasets. Notably, for the large-scale dataset (\eg Questions), many recent methods encounter the Out-of-Memory (OOM) issue as they demand the dense adjacency matrix. But our method uses the sparsified matrix and achieves competent results on Questions dataset, demonstrating the good scalability.
\begin{table*}[ht]
\centering
\caption{Ablation results (\%) of MHetGL.}
\resizebox{\linewidth}{!}{
\begin{tabular}{c|c||c|c|c|c||c|c|c|c|c|c}
\toprule
\textbf{Method} & \textbf{Metric} & \textbf{Cora} & \textbf{CiteSeer} & \textbf{ML} & \textbf{PubMed} & \textbf{Reddit} & \textbf{Weibo} & \textbf{Books} & \textbf{Disney} & \textbf{Enron} & \textbf{Questions}\\
\hline \hline

\multirow{2}{*}{\textbf{MHetGL$^{loc}$}}
& \textbf{AUPR} & \textbf{97.60} & 79.85 & 91.96 & \underline{\textsl{94.04}}
& 3.86 & 64.34 & 3.29 & 6.83 & 0.10 & \underline{\textsl{5.16}}\\
& \textbf{AUROC} & 98.32 & 99.37 & 98.02 & 98.88
& 54.06 & 89.76 & 53.82 & 36.00 & 63.80 & 61.11\\
\hline

\multirow{2}{*}{\textbf{MHetGL$^{glo}$}}
& \textbf{AUPR} & 96.14 & 99.10 & 90.52 & \textbf{96.14}
& 4.15 & \underline{\textsl{79.55}} & 7.07 & 7.08 & 0.13 & 4.32\\
& \textbf{AUROC} & 99.01 & 99.96 & 99.65 & 98.36
& 55.57 & \underline{\textsl{96.75}} & 62.74 & 39.29 & 70.81 & 60.03\\
\hline

\multirow{2}{*}{\textbf{MHetGL$_{w/o\ reg}$}}
& \textbf{AUPR} & 96.36 & \underline{\textsl{99.22}} & \underline{\textsl{90.71}} & 92.90
& \underline{\textsl{4.73}} & 81.17 & \underline{\textsl{8.20}} & 11.53 & \underline{\textsl{0.16}} & 4.41\\
& \textbf{AUROC} & \underline{\textsl{99.42}} & \underline{\textsl{99.97}} & \underline{\textsl{99.69}} & \underline{\textsl{99.08}}
& \underline{\textsl{60.27}} & 96.67 & \underline{\textsl{76.71}} & 64.14 & \underline{\textsl{76.92}} & \underline{\textsl{60.32}}\\
\hline \hline

\multirow{2}{*}{\textbf{MHetGL$^{pur}$}}
& \textbf{AUPR} & 7.51 & 89.00 & 24.37 & 63.74
& 2.59 & 77.41 & 2.87 & \underline{\textsl{27.65}} & 0.13 & 3.88\\
& \textbf{AUROC} & 59.91 & 99.50 & 75.51 & 84.97
& 39.97 & 95.69 & 56.66 & \underline{\textsl{75.57}} & 71.17 & 53.68\\
\hline

\multirow{2}{*}{\textbf{MHetGL$^{aug}$}}
& \textbf{AUPR} & 92.79 & 92.30 & 90.68 & 93.22
& 4.70 & 21.90 & 4.51 & 5.47 & 0.08 & 4.84\\
& \textbf{AUROC} & 99.09 & 99.87 & 97.99 & 98.57
& 55.20 & 79.32 & 45.47 & 27.57 & 53.41 & 57.55\\
\hline

\multirow{2}{*}{\textbf{MHetGL}} 
& \textbf{AUPR} & \underline{\textsl{96.67}} & \textbf{99.22} & \textbf{90.74} & 92.87
& \textbf{5.09} & \textbf{81.62} & \textbf{8.35} & \textbf{33.33} & \textbf{0.16} & \textbf{5.26}\\
& \textbf{AUROC} & \textbf{99.51} & \textbf{99.97} & \textbf{99.69} & \textbf{99.15}
& \textbf{65.37} & \textbf{97.50} & \textbf{77.06} & \textbf{85.00} & \textbf{76.93} & \textbf{63.03}\\
\bottomrule
\end{tabular}
}
\label{table:ablation}
\end{table*}

\subsection{Ablation Study}
Then, we investigate the impact of different components of the proposed method, as reported in Table \ref{table:ablation}. To study the effectiveness of MHL, we design three variants, where MHetGL$^{loc}$ only applies the local hypersphere learning, MHetGL$^{glo}$ only applies global hypersphere learning, and MHetGL$_{w/o\ reg}$ removes the hypersphere regularization and calculates $L^{clu}$ by Equation~\eqref{HL_contrastive} instead of Equation~\eqref{HL_contrastive_final} in the loss function. To study the effectiveness of HGE, we additionally design two variants based on our MHetGL, including MHetGL$^{pur}$ which considers only existing neighbor discrimination, and MHetGL$^{aug}$ which considers only latent neighbor discovery.

We observe the full MHetGL achieves the best performance on both the injected and organic datasets, which verifies the effectiveness of our framework for unsupervised GAD. 
MHetGL significantly outperforms MHetGL$^{glo}$ and MHetGL$^{loc}$, especially on organic datasets. These findings validate that the global and local graph contexts are complementary and should be jointly applied to GAD problem. For specific organic datasets (\eg Reddit, Disney, Questions), MHetGL$_{w/o\ reg}$ witnesses the notable performance degradation, demonstrating the significance of hypersphere regularization. It can alleviate hypersphere collapse and be more robust to hypersphere shrinking. 
Moreover, the results of MHetGL$^{pur}$ and MHetGL$^{aug}$ are unstable. However, MHetGL outperforms these two variants and behaves stably in all datasets, verifying that purifying and augmenting anomalous neighbors are complementary and significant for GAD.

\begin{table*}[h]
\centering
\caption{Negative weights.}
\resizebox{\linewidth}{!}{
\begin{tabular}{c|c||c|c|c|c||c|c|c|c|c|c}
\toprule
\textbf{Method} & \textbf{Metric} & \textbf{Cora} & \textbf{CiteSeer} & \textbf{ML} & \textbf{PubMed} & \textbf{Reddit} & \textbf{Weibo} & \textbf{Books} & \textbf{Disney} & \textbf{Enron} & \textbf{Questions}\\
\hline \hline

\multirow{2}{*}{\textbf{MHetGL$^{neg}$}} 
& \textbf{AUPR} & 37.47 & 85.90 & 66.77 & 43.86 
& 4.31 & 16.82 & 7.26 & 28.02 & 0.11 & 4.63\\
& \textbf{AUROC} & 78.15 & 97.21 & 89.11 & 94.85 
& 56.72 & 72.92 & 59.49 & 71.67 & 52.52 & 58.68\\
\hline

\multirow{2}{*}{\textbf{MHetGL}} 
& \textbf{AUPR} & \textbf{96.67} & \textbf{99.22} & \textbf{90.74} & \textbf{92.87}
& \textbf{5.09} & \textbf{81.62} & \textbf{8.35} & \textbf{33.33} & \textbf{0.16} & \textbf{5.26}\\
& \textbf{AUROC} & \textbf{99.51} & \textbf{99.97} & \textbf{99.69} & \textbf{99.15}
& \textbf{65.37} & \textbf{97.50} & \textbf{77.06} & \textbf{85.00} & \textbf{76.93} & \textbf{63.03}\\

\bottomrule
\end{tabular}
}
\label{table:neg_weight}
\end{table*}

\subsection{Impact of Negative Edge Weight}\label{section:neg_weight}

To study the impact of negative edge weights, we propose a variant MHetGL$^{neg}$ which omits the Softmax normalization in Equation~\eqref{discriminate_weight}.
As reported in Table~\ref{table:neg_weight}, using negative weights severely affects the performance. A possible reason is that most links in the graph represent homophilic information and they are assigned with negative curvature weights. This will hinder stable training and degrade performance.
\subsection{Parameter Analysis}\label{section:parameter}
We further study the parameter sensitivity of MHetGL, including two loss weights, the hidden dimension, the number of clusters and the derivation of the center of the global hypersphere. Due to space limitation, we present the results for part of the datasets: Cora, CiteSeer, and PubMed. The results for other datasets are similar. 

\begin{figure}[h]
    \centering
    \subfigure[Cora.]{\includegraphics[width=0.24\linewidth]{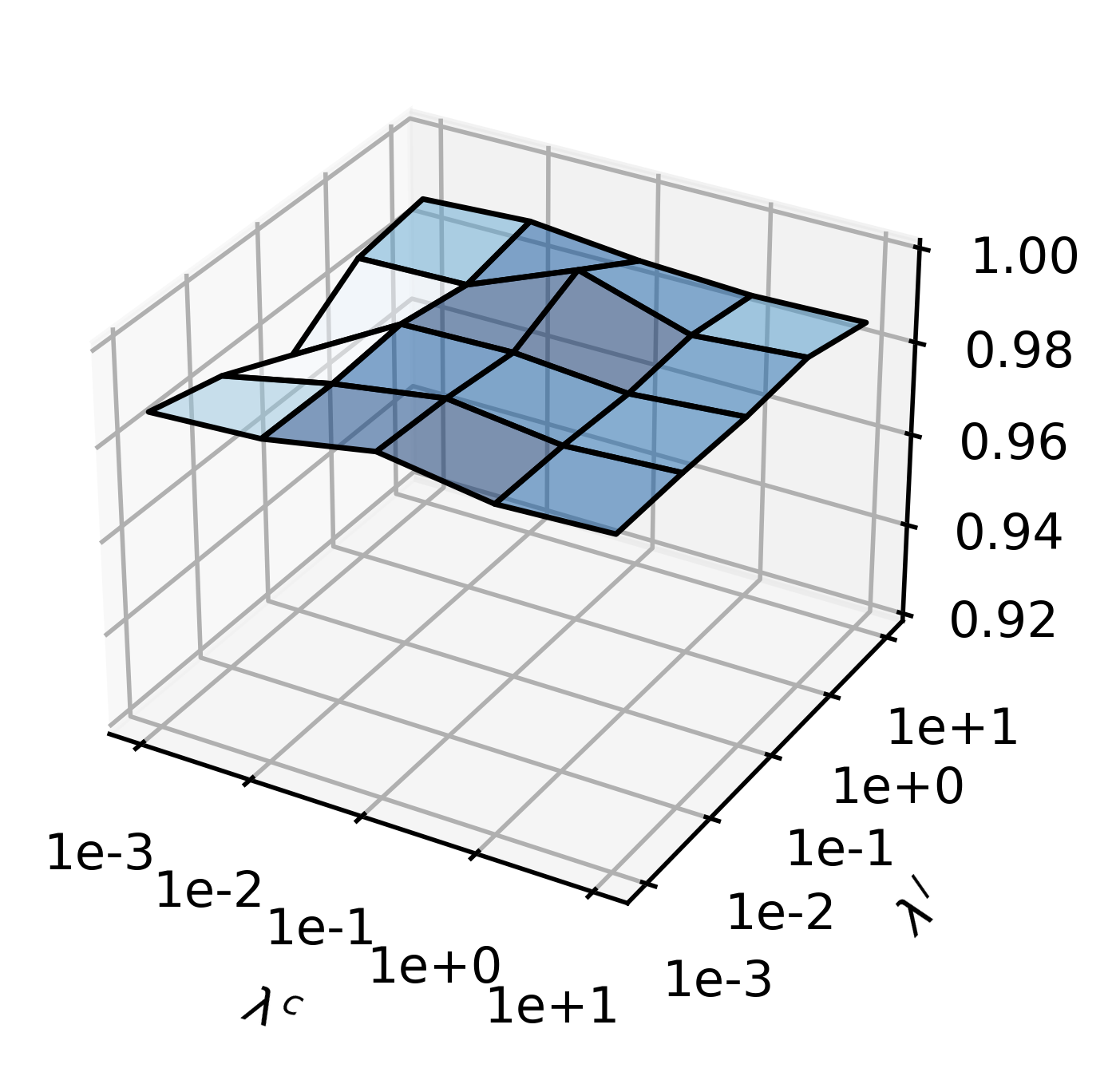}}
    \subfigure[CiteSeer.]{\includegraphics[width=0.24\linewidth]{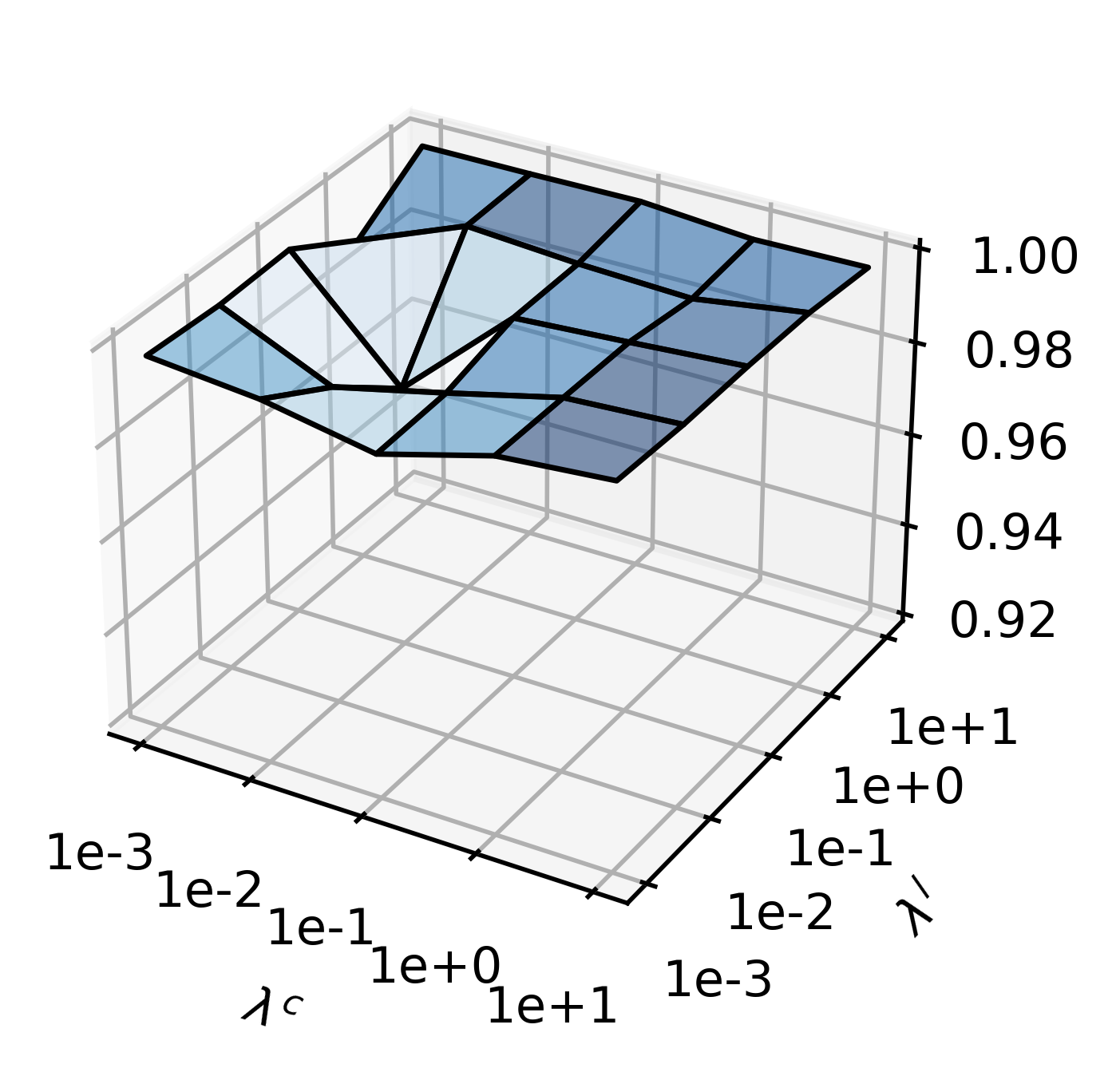}}
    \subfigure[PubMed.]{\includegraphics[width=0.24\linewidth]{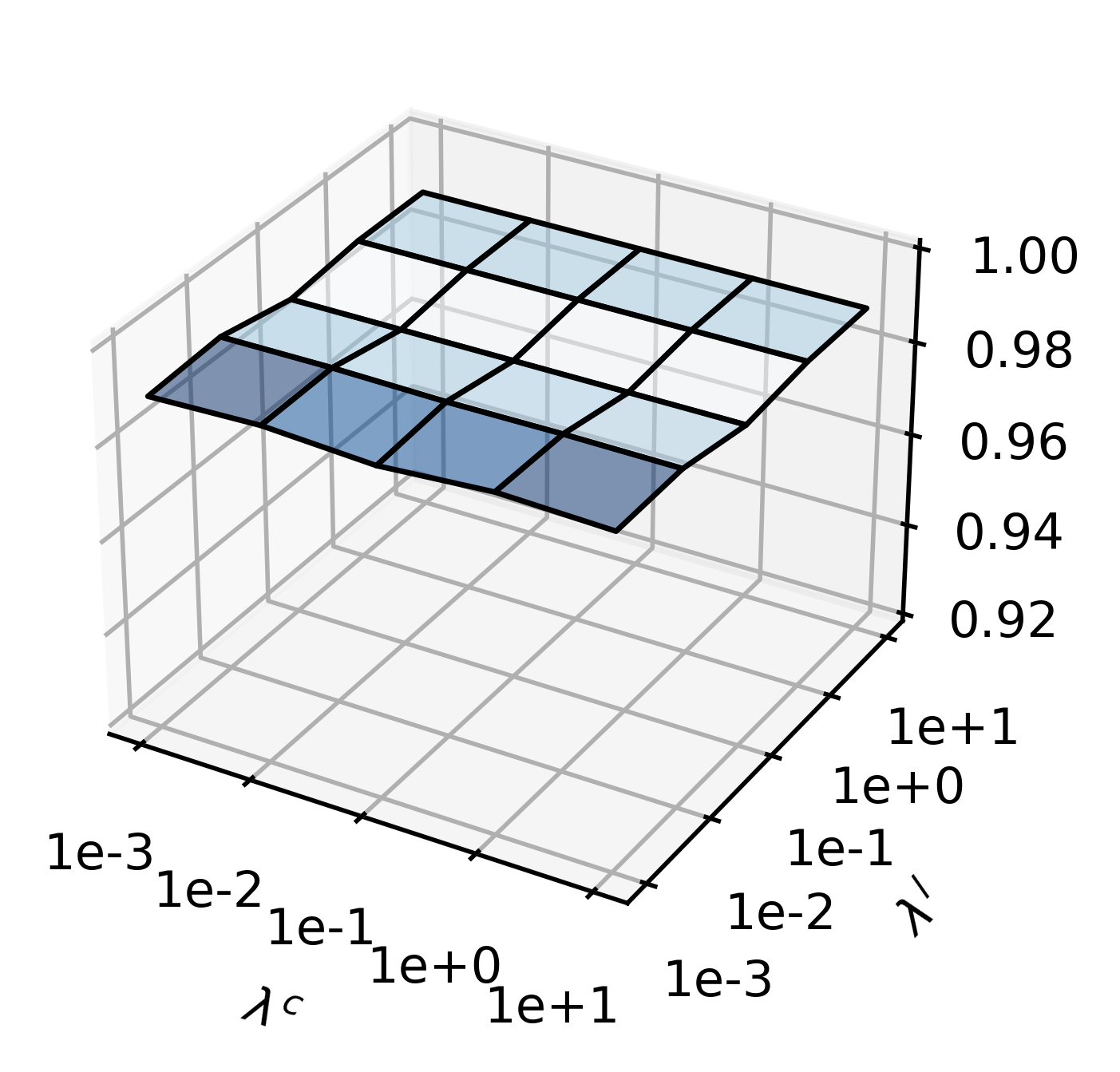}}
    \caption{Parameter sensitivity of loss weights.}
    \label{figure:loss_weights_joint}
\end{figure}

\begin{figure*}[h]
    \centering
    \subfigure[Effect of $\lambda^l$.]{\includegraphics[width=0.3\linewidth]{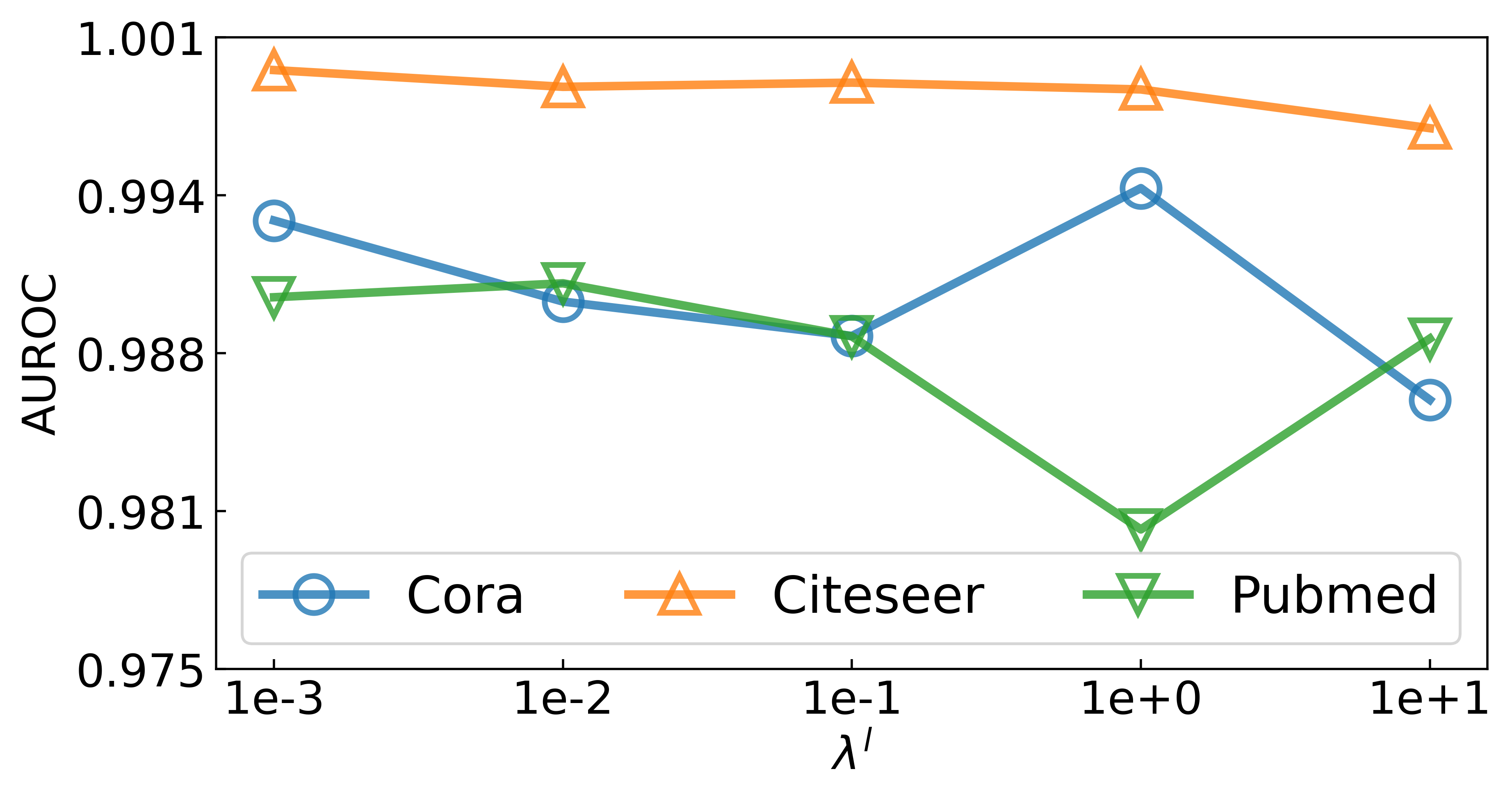}\label{figure:lambda_local}}
    \subfigure[Effect of $\lambda^c$.]{\includegraphics[width=0.3\linewidth]{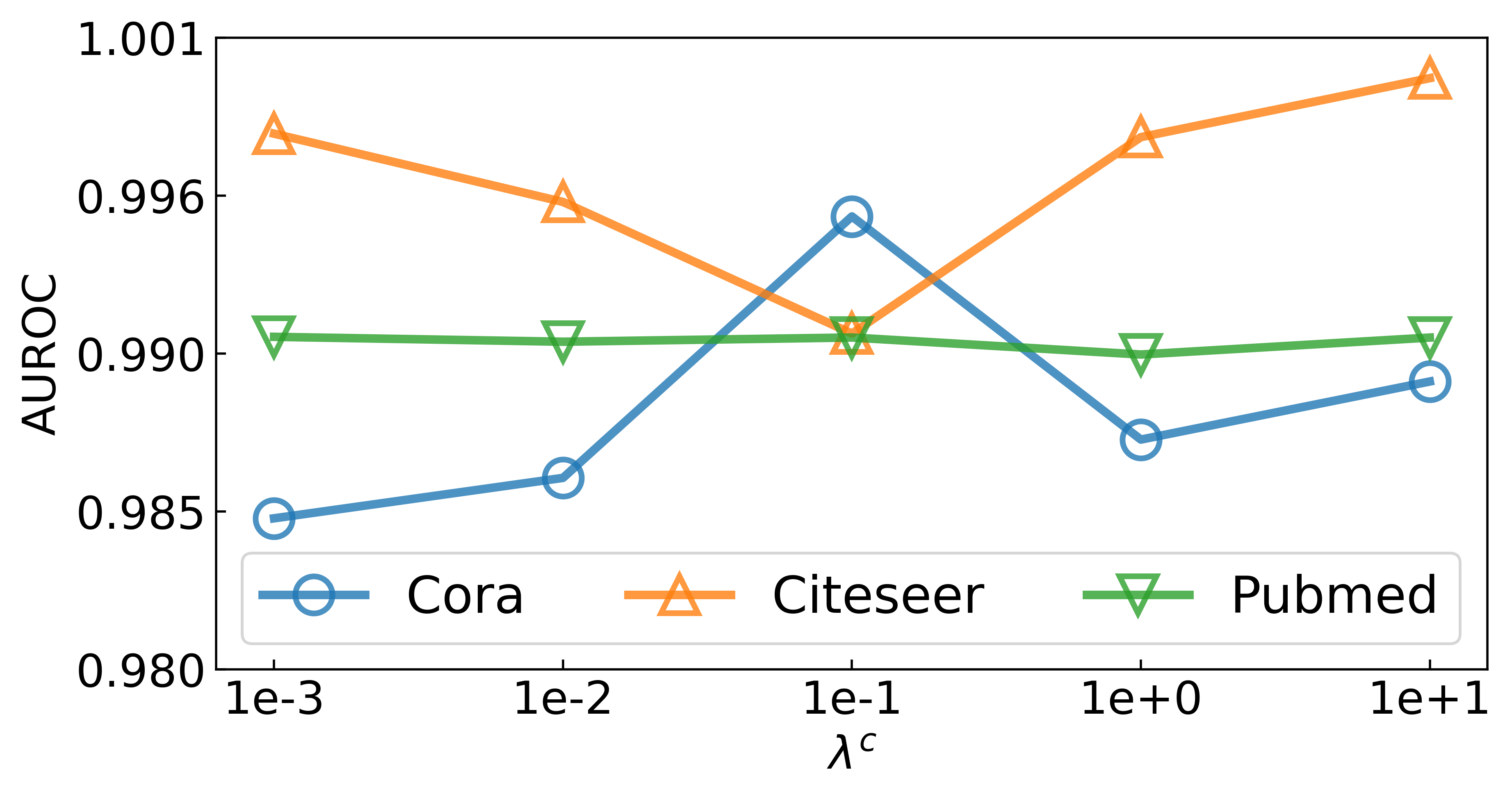}\label{figure:lambda_cluster}}\\
    \vspace{1pt}
    \centering
    \subfigure[Effect of $d$.]{\includegraphics[width=0.3\linewidth]{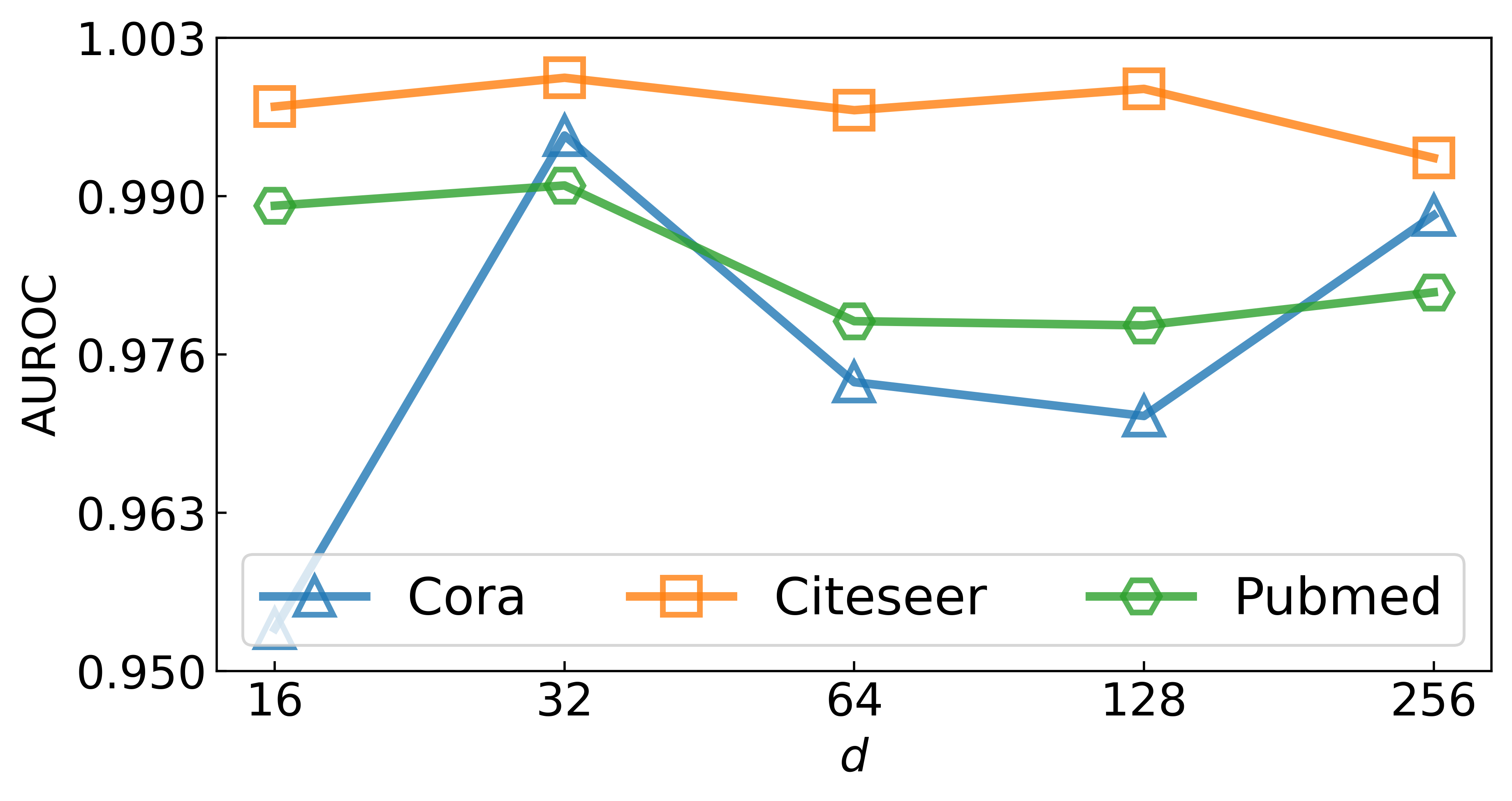}\label{figure:hidden_dim}}
    \subfigure[Effect of $K$.]{\includegraphics[width=0.3\linewidth]{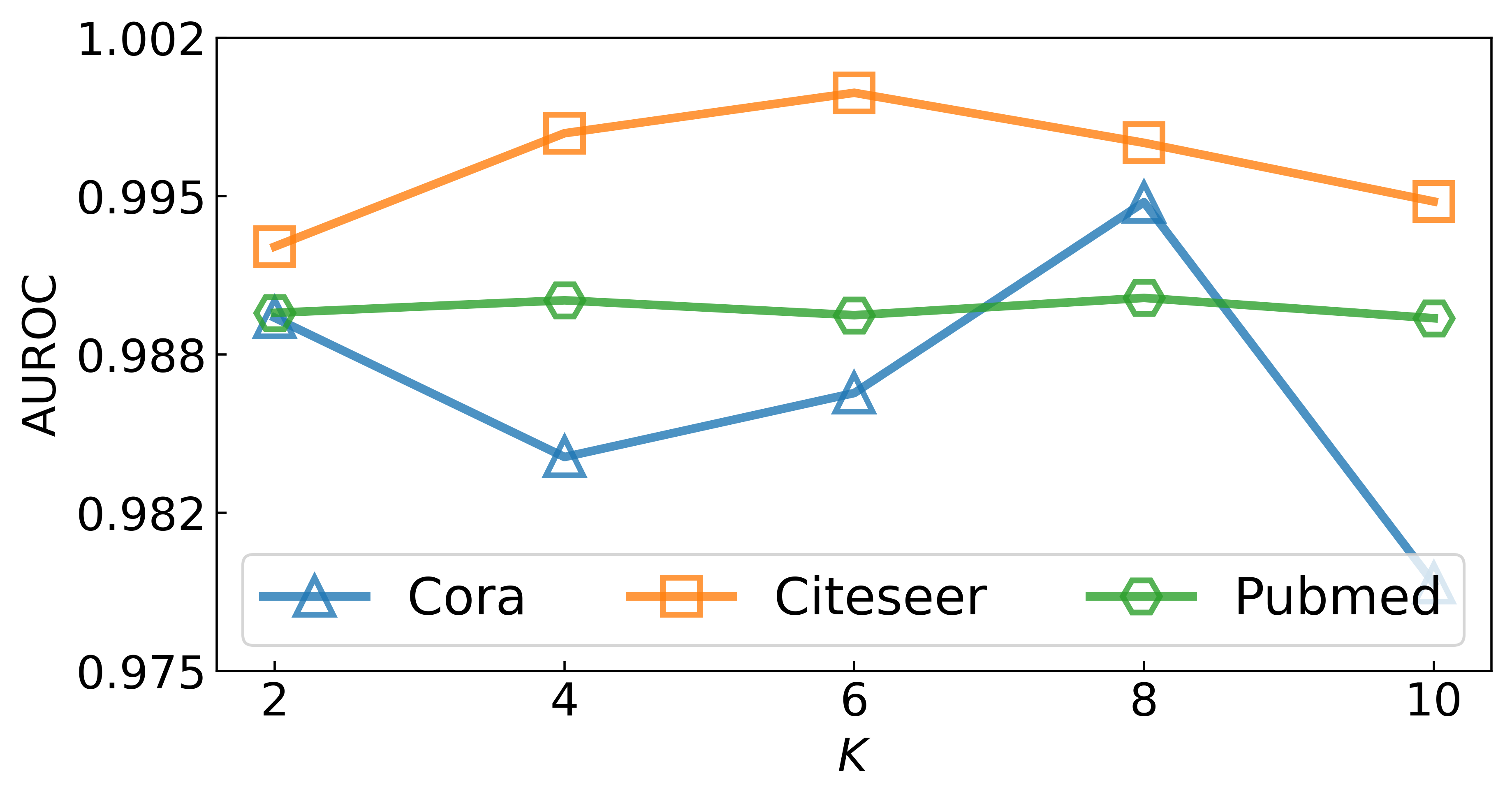}\label{figure:num_cluster}}
    \caption{Parameter sensitivity.}
    \label{figure:parameter}
\end{figure*}

\noindent\textbf{Loss weights.} We vary the loss weights $\lambda^{loc}$ and $\lambda^{clu}$ respectively from $0.001$ to $10$ and report the AUROC results of three injected datasets in Figure~\ref{figure:lambda_local} and Figure~\ref{figure:lambda_cluster}. We observe that the results in three datasets are stable and can keep in a high range. 
Further, we present a joint analysis of these two loss weights in Figure~\ref{figure:loss_weights_joint}. We vary the loss weights $\lambda^l$ and $\lambda^c$ respectively from $0.001$ to $10$ and report the AUROC results of Cora, CiteSeer and PubMed datasets in Figure~\ref{figure:loss_weights_joint}. We observe that the results in three datasets are stable and maintain a high level. The joint analysis of these two loss weights demonstrates that our method is insensitive to the loss weights and holds a stable performance.

\noindent\textbf{Hidden dimension.} We vary the hidden dimension $d$ from $16$ to $256$ and report the AUROC results in Figure~\ref{figure:hidden_dim}. We observe that, as the increase of $d$, the AUROC first increases and then drops. A possible reason lies in the  over-fitting problem caused by explosive parameters and the curse of dimensionality.

\noindent\textbf{Number of clusters.} We vary the number of clusters $K$ from $2$ to $10$ and report the AUROC in Figure~\ref{figure:num_cluster}. We observe the performance firstly increase and then decrease when we increase $K$. The AUROC value drops when $K$ is large, which indicates that setting too many communities may bring more noise and hurt performance.

\begin{figure}[h]
\centering
\includegraphics[width=0.5\linewidth]{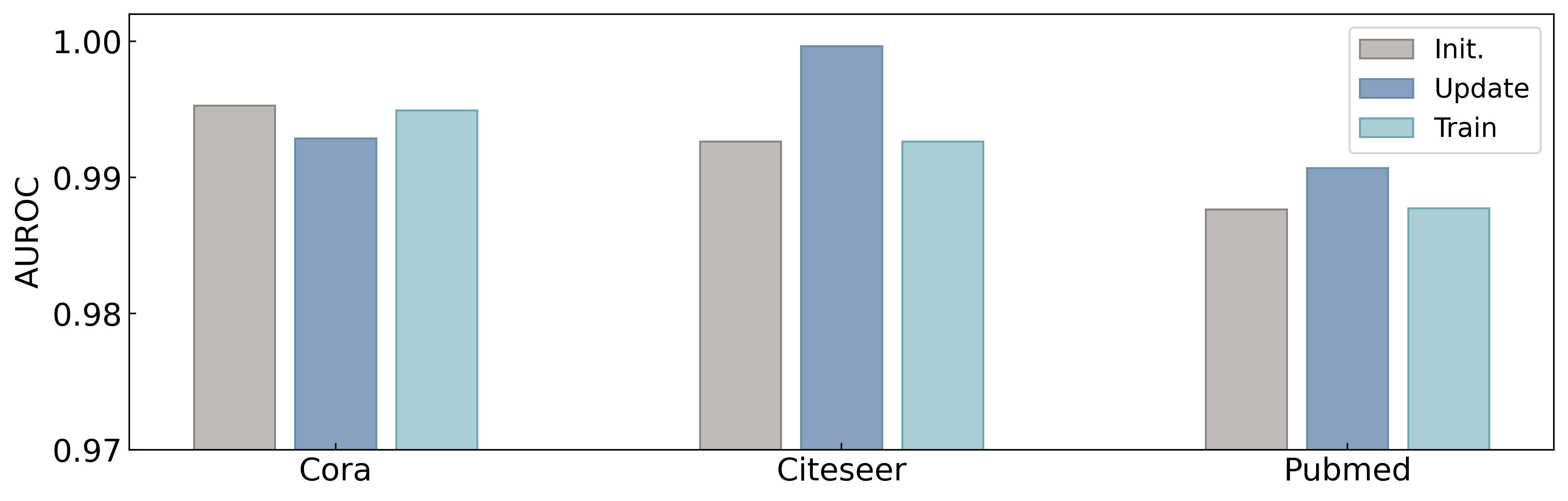}
\caption{The effect of center derivation.}
\label{figure:center_mode}
\end{figure}

\noindent\textbf{Center Derivation} 
To derive the global hypersphere center $\mathbf{c}_0$, we propose three simple yet effective strategies: (1) \textsl{Init.}: The center is computed by averaging all the node representations obtained from an initialized encoder and remains unchanged. (2) \textsl{Update}: After each epoch, we recalculate the center by averaging all the updated representations. (3) \textsl{Train}: We regard the center as a learnable vector and optimize it during model training.
We vary the options for $\mathbf{c}_0$ and report the AUROC results of Cora, CiteSeer, and PubMed datasets in Figure~\ref{figure:center_mode}. We observe that the results of the \textsl{Init.} and \textsl{Train} strategies are similar, but the \textsl{Update} strategy achieves a distinct performance. Therefore, it is difficult for neural networks to learn a better center than an initialized and fixed one, indicating that our method is insensitive to the center initialization. However, updating the center per epoch may adjust the position of the center timely and affect the final performance.
\begin{figure}[ht]
    \centering
    \subfigure[Effectiveness of the HGE module.]{\includegraphics[width=0.4\linewidth]{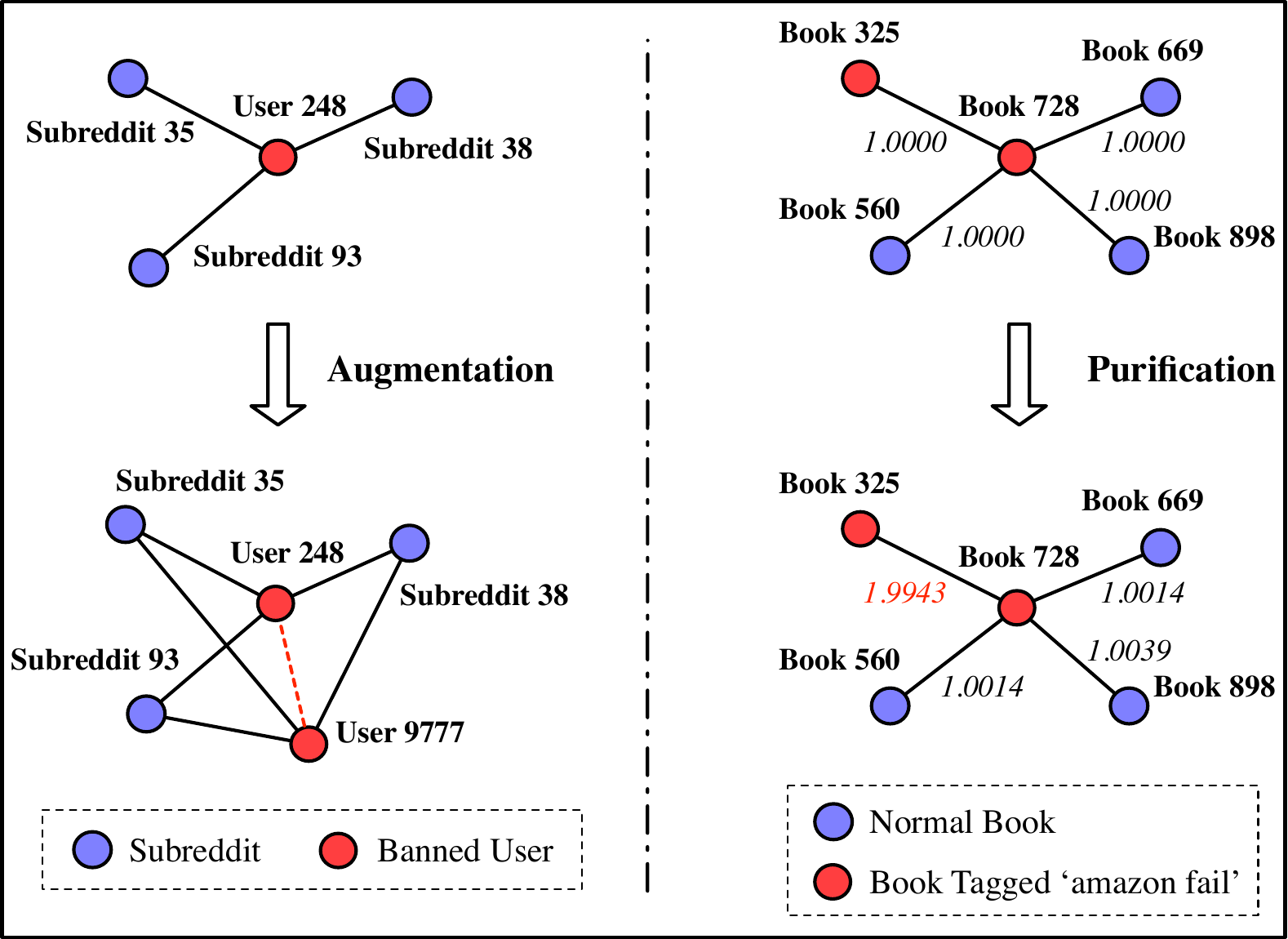}\label{figure:case_study_1}}\\
    \vspace{1pt}
    \centering
    \subfigure[Effectiveness of the MHL module.]
    {\includegraphics[width=0.4\linewidth]{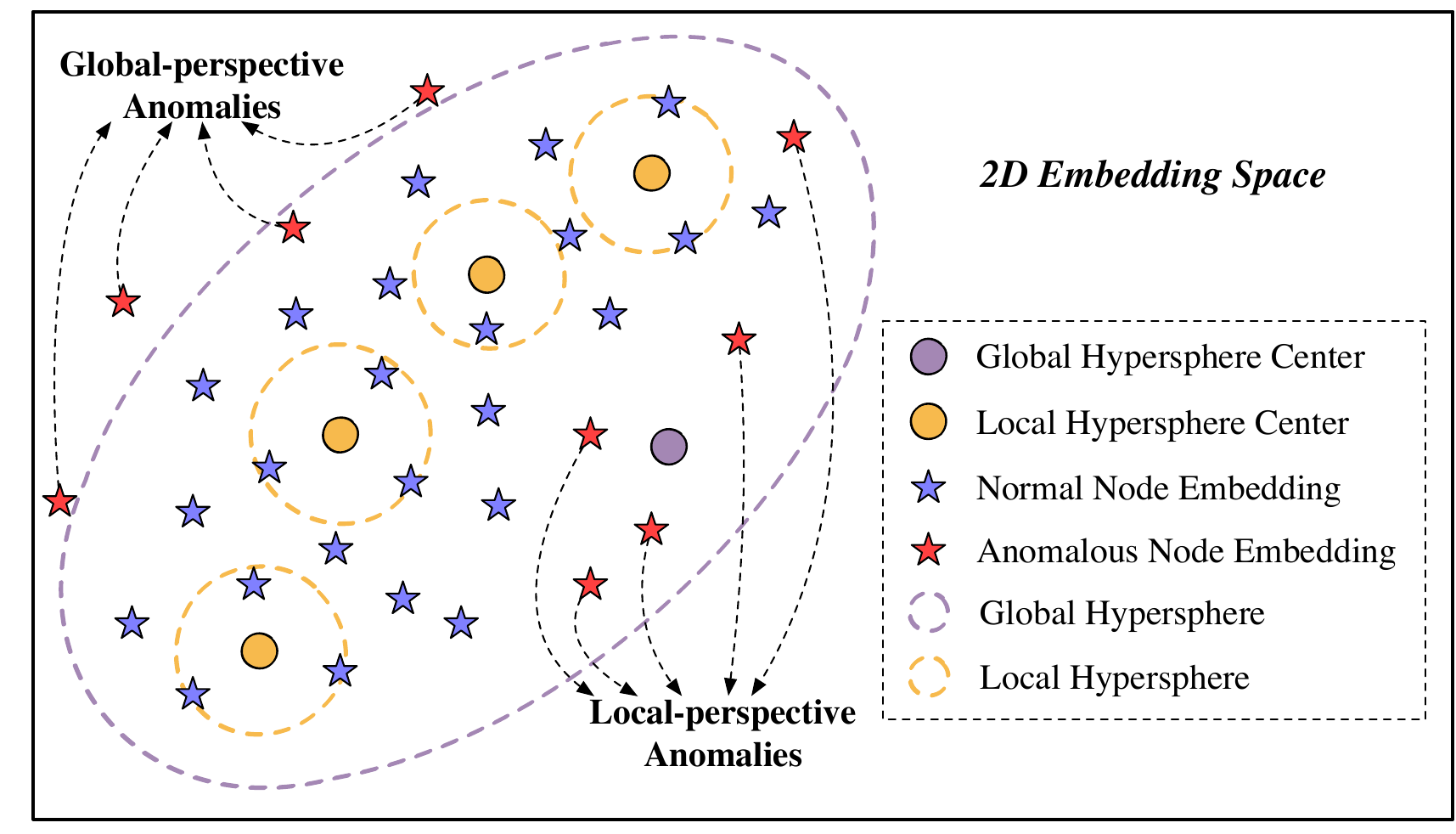}\label{figure:case_study_2}}
    \caption{The case study of the MHetGL model.}
    \label{figure:case_study}
\end{figure}

\subsection{Case Study}\label{section:case_study}
In this section, we present some cases in Figure~\ref{figure:case_study} to illustrate our motivations.
In Figure~\ref{figure:case_study_1}, we demonstrate our HGE module could refine the neighborhood of anomalies by purification and augmentation blocks, in order to conquer the homophily-induced indistinguishability. 
Specifically, we visualize the edge weights of book node indexed by $728$ and its neighbors in Books dataset. We observe that the anomalous-anomalous connection between books is enhanced by the proposed curvature-based purification block. 
In addition, we select the user node indexed by $248$ in Reddit dataset (which is a bipartite network with users and subreddits) for visualization and we can see that the target anomalous user could be linked with another anomalous user via our GDV-based augmentation block. 
In Figure~\ref{figure:case_study_2}, we show our MHL module could generate local perspectives beyond the global hypersphere, to address the uniformity-induced indistinguishability. 
Concretely, we choose the Books dataset as an example and visualize partial node embeddings in a 2D plot. Note that the hypersphere may degenerate into an ellipsoid after dimensionality reduction. We find that several anomalies are located around the global hypersphere center in the embedding space, which are difficult to discover with the vanilla hypersphere learning method. However, in our MHL module, we generate several local hypersphere centers, which could easily identify these local-perspective anomalies with proximity measures.
\section{Conclusion}
This paper presented a two-stage framework MHetGL for unsupervised GAD.
Specifically, we proposed a Heterophilic Graph Encoding (HGE) module to learn discriminative node representations.
In particular, HGE first manipulates the graph topology to enhance the graph homophily for anomalous nodes and then aggregates neighbor information by conducting message passing on the manipulated graph structure.
Moreover, we construct a multi-hypersphere learning module to enhance context-dependent anomaly distinguishability.
In particular, HGE devises multiple global and local hyperspheres for collective anomaly identification, and a tailored hypersphere regularization block to avoid trivial solutions in multi-hypersphere learning. 
Extensive experimental results demonstrated that MHetGL achieves consistent state-of-the-art performance compared to 14 unsupervised GAD baselines on ten real-world datasets.


\bibliographystyle{ACM-Reference-Format}
\bibliography{ref}

\end{document}